\documentclass{article}

\usepackage[margin=1in]{geometry}
\usepackage{graphicx}
\graphicspath{{images/}}
\usepackage{subfigure}
\usepackage{multirow}
\usepackage{array}
\usepackage{amssymb}
\usepackage{amsmath}
\usepackage[font=small, labelfont=bf, skip=6pt]{caption}

\usepackage[colorlinks=true,linkcolor=blue,citecolor=blue,urlcolor=blue]{hyperref}

\title{\textbf{Pillar-Voxel Fusion Network for 3D Object Detection in Airborne Hyperspectral Point Clouds}}

\author{
  Yanze Jiang \quad Yanfeng Gu \quad Xian Li\thanks{Corresponding author: xianli@hit.edu.cn} \\
  School of Electronics and Information Engineering,\\
  Harbin Institute of Technology, Harbin 150001, China
}

\date{}  

\begin{document}

\maketitle

\begin{abstract}
{Hyperspectral point clouds (HPCs) can simultaneously characterize 3D spatial and spectral information of ground objects, offering excellent 3D perception and target recognition capabilities. Current approaches for generating HPCs often involve fusion techniques with hyperspectral images and LiDAR point clouds, which inevitably lead to geometric-spectral distortions due to fusion errors and obstacle occlusions. These adverse effects limit their performance in downstream fine-grained tasks across multiple scenarios, particularly in airborne applications. To address these issues, we propose PiV-AHPC, a 3D object detection network for airborne HPCs. To the best of our knowledge, this is the first attempt at this HPCs task. Specifically, we first develop a pillar-voxel dual-branch encoder, where the former captures spectral and vertical structural features from HPCs to overcome spectral distortion, while the latter emphasizes extracting accurate 3D spatial features from point clouds. A multi-level feature fusion mechanism is devised to enhance information interaction between the two branches, achieving neighborhood feature alignment and channel-adaptive selection, thereby organically integrating heterogeneous features and mitigating geometric distortion. Extensive experiments on two airborne HPCs datasets demonstrate that PiV-AHPC possesses state-of-the-art detection performance and high generalization capability.}
\end{abstract}

\vspace{0.5em}
\noindent\textbf{Keywords:} Hyperspectral Point Cloud, 3D Object Detection, Feature Fusion, Geometric-Spectral Distortion

\section{Introduction}
Hyperspectral Point Clouds (HPCs) acquired from airborne platforms integrate the advantages of LiDAR point clouds and hyperspectral imaging, providing rich spectral and spatial information\cite{1,2,3,4}. By integrating precise 3D measurements with detailed spectral signatures, HPCs enable accurate inversion of both material properties and spatial structures of surface objects. This unique combination of capabilities demonstrates significant potential across multiple applications, including 3D urban planning, precision agriculture, forest management, and camouflage detection\cite{5,6,7,8,9,10}.

HPCs generation and data processing is the premise of its practical application. In the last decade, several fusion methods have been explored to generate HPCs from sub-pixel to pixel to super-voxel. Brell et al. \cite{12} proposed a segmentation-based spatial unmixing method to generate sub-pixel HPCs with airborne point clouds and hyperspectral images. Subsequently, Fu et al. \cite{13} designed a deep learning method to learn the similarity between hyperspectral images and point clouds, achieving pixel-level HPCs generation. Lately, Xie et al. \cite{14} yielded super-voxel-level shadow-free HPCs with super-voxel segmentation and shadow removal via global illumination estimation. Regarding HPCs data processing, several segmentation models have been applied to different acquisition platforms and application scenes. Chen et al. \cite{15} designed a three-stage segmentation algorithm for indoor HPCs, which performs better than point clouds due to using both geometric and spectral information. Mitschke et al. \cite{16} extended RandLA \cite{17} to the ground-based HPCs, improving the segmentation performance by the use of spectral information. Afifi et al. \cite{18} analyzed the effectiveness of several networks on the outcrop HPCs. Recently, Fu et al. \cite{13}, building on high-precision algorithms \cite{19,20}, achieved superior 3D mapping of karst wetland vegetation on airborne HPCs compared to 2D methods.


Although the above-mentioned studies demonstrate that HPCs have great potential in many applications, there is currently no research on 3D object detection tasks. Numerous 3D object detection models \cite{21,22,23,24,25,26,27,28,29,54,55} were designed for LiDAR since it directly obtains accurate 3D spatial information of objects, and operates continuously all-time. LiDAR-based 3D object detection networks primarily consist of two key components: an encoder and a detection head. The encoder is responsible for extracting data features, while the detection head generates detection proposals based on the extracted features. Depending on the input data format, encoders can be categorized into three types: point-based, voxel-based, and pillar-based. Point-based encoders \cite{30,31,32,33} employ point-wise feature extraction and multi-stage optimization strategies, which fully preserve the geometric information of point clouds but incur high computational costs. To reduce computational complexity, researchers have proposed converting point clouds into standardized formats, such as voxels or pillar units. Voxel-based encoders \cite{34,35,36} have evolved from initial dense convolutions \cite{56} to sparse convolutions \cite{34} that leverage voxel sparsity, and further to innovative designs like point-voxel structures \cite{37,38} that combine the advantages of point clouds and voxels, as well as dynamic sparse windows that build local voxel associations. In contrast, pillar-based encoders \cite{39,40,41,42} typically offer higher computational efficiency. By optimizing the encoder structure or introducing feature pyramids, they can improve detection accuracy while maintaining high time efficiency.

As another critical component, the detection head follows several mainstream design strategies, including center-based, refinement-based, query-based, and sparse prediction-based methods. Center-based methods \cite{43,44} use Gaussian heatmaps to predict target center locations, replacing traditional anchor box schemes, thereby reducing the search space and improving algorithm efficiency. Refinement-based \cite{35,58} methods introduce a second-stage network to further optimize the category and size of detection boxes based on initial predictions. Query-based methods \cite{46,60} draw inspiration from the Transformer decoder mechanism, using query vectors as target representations and matching them with corresponding spatial features to generate detection results. Sparse prediction-based methods \cite{47,48,49,57,59} abandon dense bird's-eye view (BEV) representations and instead use instance voting or voxel features as proxies for detection proposals, demonstrating superior computational efficiency in large-scale scenes.

\begin{figure}[h]
    \centering
    \includegraphics[width=0.42\textwidth,height=0.3\textwidth]{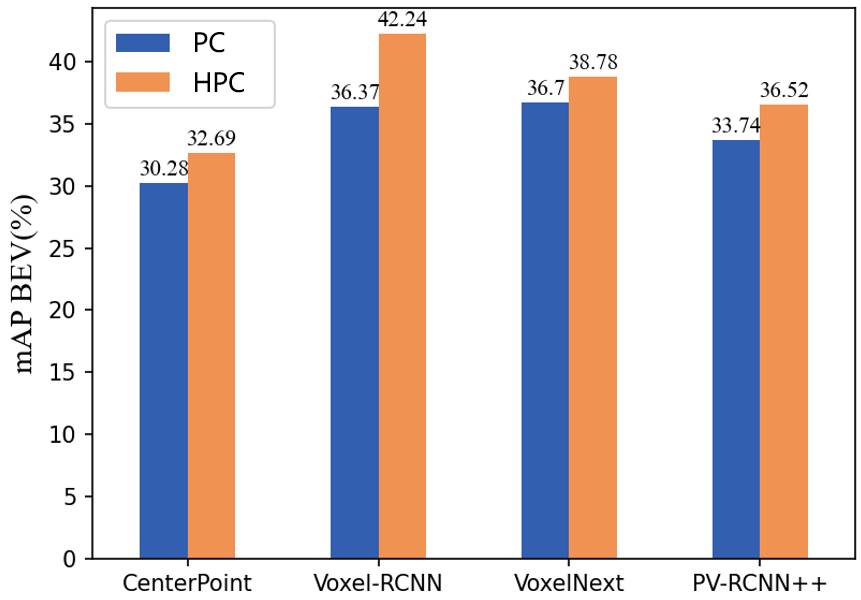}
    \caption{Detection performance of four 3D object detection networks over point cloud and HPCs in HIT Campus Dataset.}
    \end{figure} 

A limitation of 3D object detection for LiDAR point clouds is a weaker detection capability due to the lack of spectral signatures, especially for airborne scenes with the interference of structural similar background. A clever idea is to attempt to exploit airborne HPCs that contain 3D spatial and rich spectral information, which holds great potential for accurate 3D object detection, as shown in Figure 1. It is also evident that the gains of current object detection networks in mean average precision (mAP) are limited. The essential reasons for this phenomenon include: 

\begin{itemize}
  \item [1)] 
  \textbf{High Spatial-Spectral Complexity: } Due to the diversity of spatial distribution and the variability of spectral signatures, airborne HPCs often have high spatial-spectral complexity between different objects and backgrounds.     
  \item [2)]
  \textbf{Spectral Distortion: } Due to the inability of passive hyperspectral imaging to penetrate obstacles (e.g., tree canopies) and the resolution differences between hyperspectral images and point clouds, occluded objects and edges inevitably exhibit spectral distortion, as shown in Figure 2(a).
  \item [3)]
  \textbf{Geometric Distortion: } Due to different imaging patterns between LiDAR and hyperspectral sensors, generated HPCs with fusion techniques occur geometric distortion in shape, size, and position, as shown in Figure 2(b).
\end{itemize}

Inspired by these challenges, we propose a Pillar-Voxel Fusion Network for 3D Object Detection in Airborne Hyperspectral Point Clouds, named PiV-AHPC. We design an intuitive and effective pillar-voxel dual-branch encoder to extract complex heterogeneous features and overcome spectral distortion. Due to the advantage of the pillar structure's vertical receptive field and its natural adaptability to the BEV, this branch extracts spectral features for classification and determines vertical structure features as spectral confidence criteria. Simultaneously, since the voxel structure preserves the spatial structure details, this branch captures fine-grained three-dimensional spatial features for precise target localization. The pillar branch extracts vertical structural features that focus on capturing point cloud distribution along the z-axis, while the voxel branch extracts 3D spatial features that emphasize capturing fine 3D geometric details. We implement a multi-level feature fusion mechanism that contains two modules to mitigate geometric distortion and enhance branch correlation. Specifically, intermediate-level branch correlations are established by the sparse fusion module to provide more information for subsequent bounding box refinement. During the output of the BEV feature map, the patch-wise fusion module achieves an organic integration of neighboring features through adaptive selection and alignment. This well-designed architecture ensures the synergy of the encoders and fully mines the rich information in airborne HPCs.


\begin{figure}[h]
    \centering
    \includegraphics[width=0.5\textwidth]{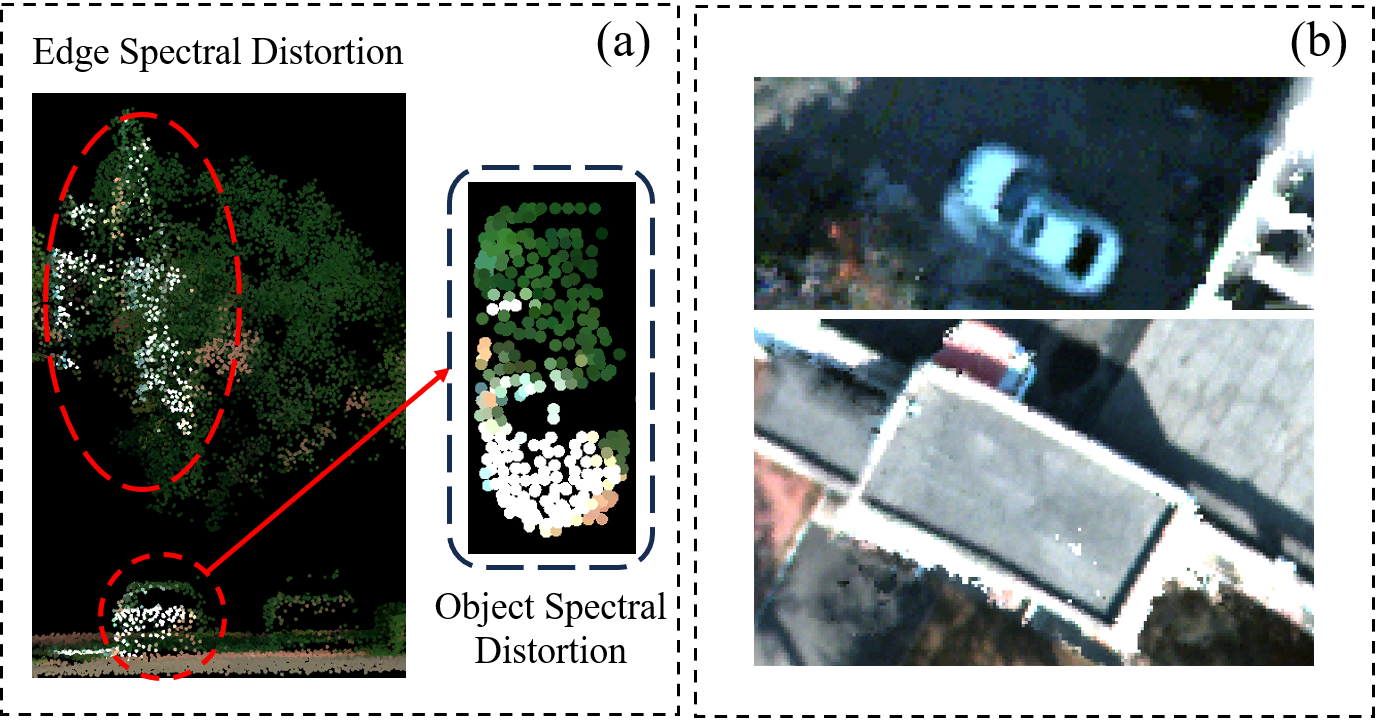}
    \caption{Problem visualization for HPCs. (a) Spectral distortion. (b) Geometric distortion}
    \end{figure}

The main contributions of our work are given as follows:
\begin{itemize}
  \item [1)] 
We propose a novel 3D object detection framework for airborne HPCs. To the best of our knowledge, there are no reported works in this area. The main advantage of the proposed is its discriminative 3D spatial-spectral representation capability, which demonstrates robust performance in complex airborne scenes with occluded and fake targets.   
  \item [2)]
We develop an elegant pillar-voxel dual-branch encoder that extracts spectral, vertical structural, and 3D spatial features of airborne HPCs in parallel, sharply mitigating the adverse effects of spectral distortion.
  \item [3)]
We devise a multi-level feature fusion mechanism based on sparse and patch-wise fusion strategies, which establishes feature interactions between the branches at different levels, significantly enhancing the feature mining ability. 
\end{itemize}

The rest of this article is organized as follows. Section 2 describes the proposed network structure and sub-module details. Section 3 provides numerous experiments and analyses of the detection results. Section 4 gives the conclusion of the research.

\section{Methodology}
\subsection{Overall architecture}

We propose PiV-AHPC, the first 3D object detection network specifically designed for airborne HPCs. This network captures both 3D spatial and spectral features in complex airborne scenes, demonstrating strong robustness to interference. The core components of PiV-AHPC include a pillar-voxel dual-branch encoder and multi-level feature fusion mechanism. The encoder mitigates the impact of spectral distortion through targeted extraction of spectral, vertical structure, and 3D spatial features. The fusion mechanism enhances feature extraction capability through interaction and fusion of dual-branch features, further overcoming challenges posed by geometric distortion.

First, we reduce the spectral dimensions of HPCs through Principal Component Analysis (PCA, default reduction to 21 components), and input the pillar branch after pillarization. This PCA operation can eliminate redundant information and reduce computational resource consumption. Simultaneously, we split point clouds containing positional information from the HPCs and input the voxel branch after voxelization. During the feature encoding process in both branches, the middle-layer features are dynamically adjusted and integrated by a sparse feature fusion module. At the end of the two encoders, a patch-wise feature fusion module adaptively aligns and selects dual-branch neighborhood feature channels, obtaining a BEV representation of the scene information. Finally, this representation is fed into a detection head with the same structure as the Voxel-RCNN to produce the prediction results. The overall architecture is illustrated in Figure 3.

\begin{figure}[h]
    
    \centering
    \includegraphics[width=1\textwidth,height=0.6\textwidth]{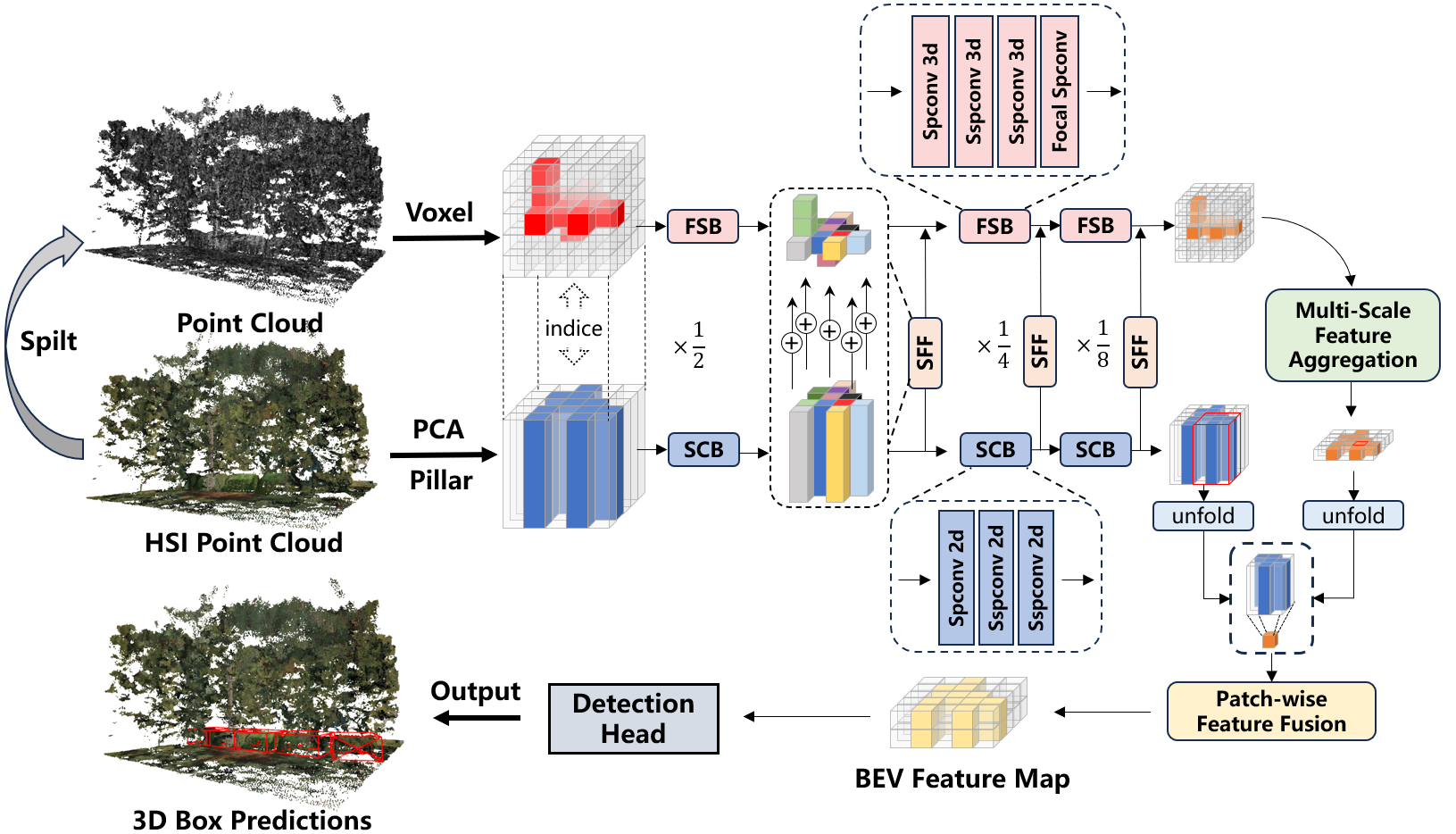}
    \caption{The overall architecture of the proposed PiV-AHPC. First, point clouds are split from HPCs and converted to voxels, while HPCs is transformed into pillars. Next, pillar and voxel features are encoded using Focus Sparse Blocks (FSB) and Sparse Convolution Blocks (SCB) respectively. Through the Sparse Feature Fusion (SFF) module, intermediate-layer features are dynamically adjusted to optimize feature representation. Subsequently, the path-wise feature fusion module performs adaptive alignment and channel selection on feature maps from both branches, generating a BEV feature map that integrates information from both paths. Finally, this feature map is input to the detection head to generate object detection proposals.}
    \end{figure} 

\subsection{Pillar-Voxel Dual-Branch Encoder}
Conventional encoders for 3D object detection typically extract features from a single data structure, such as voxels or pillars. This approach limits their ability to fully exploit the rich spatial-spectral information in airborne HPCs and to mitigate inherent spectral distortions.

Pillars can be viewed as BEV-projected representations, naturally adapting to airborne hyperspectral imagery.    In addition, due to the inability of passive hyperspectral imaging to penetrate obstacles (such as tree canopies), occluded objects and edges inevitably exhibit spectral distortion. Pillars possess sufficient receptive field in the vertical direction, allowing them to capture structural information throughout the entire vertical space, thereby identifying spectral distortion areas where occlusion exists. Finally, the excellent computational efficiency of pillars ensures the algorithm's running speed when handling redundant HPCs.

Voxels, on the other hand, divide the point cloud into regular 3D grids.    In airborne scenes characterized by high similarity and diverse distribution of ground objects,  voxels that retain more scene details than pillars could better extract fine-grained 3D spatial features and accurately locate targets.

Based on this analysis,integrating the structural advantages of both pillars and voxels is an intuitive strategy for extracting airborne HPCs features.    Therefore, we design a pillar-voxel dual-branch encoder, where the pillar branch extracts spectral and vertical structural features, while the voxel branch captures 3D spatial structural features, achieving complementary advantages. The algorithm uses the vertical structural features extracted by the pillar branch as criteria for spectral reliability, thereby evaluating the reliability of spectral data in different regions. When an area with spectral distortion is identified, the network adaptively reduces its dependence on the spectral features of that region, instead relying more on the fine-grained spatial structural features extracted by the voxel branch. This adaptive mechanism effectively mitigates the interference of spectral distortion on detection performance.


Firstly, we partition the HPCs and point clouds using an identical horizontal grid size within the designated detection range, as illustrated below:

\begin{align}\left\{\begin{aligned}
\begin{matrix}
   \left[ {{X}_{v}},{{Y}_{v}},{{Z}_{v}} \right]={{I}_{v}}\times \left[ S_{x}^{v},S_{y}^{v},S_{z}^{v} \right]  \\
   \left[ {{X}_{p}},{{Y}_{p}} \right]={{I}_{p}}\times \left[ S_{x}^{p},S_{y}^{p} \right]  \\
   \left[ S_{x}^{v},S_{y}^{v}\left] = \right[S_{x}^{p},S_{y}^{p} \right]  \\
\end{matrix} 
\end{aligned}\right.\end{align}

The parameters ${\left[ S_{x}^{v},S_{y}^{v},S_{z}^{v} \right]}$ and ${\left[ S_{x}^{p},S_{y}^{p} \right]}$ represent the sizes of voxel and pillar. The indices of the pillars and voxels are denoted as ${{I}_{p}}\in {{\mathbb{N}}^{{{N}_{p}}\times 2}}$ and ${{I}_{v}}\in {{\mathbb{N}}^{{{N}_{v}}\times 3}}$, where ${{N}_{p}}$ and ${{N}_{v}}$ are the number of non-empty pillars and voxels in their respective branches. $\left[ {{X}_{p}},{{Y}_{p}} \right]$ and $\left[ {{X}_{v}},{{Y}_{v}},{{Z}_{v}} \right]$ refer to the sets of actual coordinates for pillars on the horizontal plane and voxels in 3D space, respectively.

In the pillar branch, a lightweight PointNet encoder first processes point clouds divided into identical pillars, incorporating features such as absolute coordinates, normalized coordinates, center points, elevation range, and spectral data to obtain the initial features ${{F}_{p}}\in {{\mathbb{R}}^{{{N}_{p}}\times {{C}_{p}}}}$, where $C_p$ represents the number of feature channels in the Pillar branch. This branch consists of four stages: the first stage uses two layers of submanifold sparse convolution \cite{50} for preprocessing, while the remaining stages combine one layer of 2D sparse convolution with two layers of submanifold sparse convolution to form a 2D sparse convolution block (SCB), as shown in Figure 3. Efficient sparse convolution operators downsample the scene, reducing computational load by focusing on relevant features and ignoring redundant spatial data. Submanifold convolution maintains data sparsity and structural integrity by restricting output to existing non-empty elements. Finally, the 8$\times $downsampled sparse features $F_{p}^{4}$ are transformed into dense pillar features $F_{p}^{D}\in {{\mathbb{R}}^{H\times W\times C}}$ based on indices $I_{p}^{4}$.

In the voxel branch, the mean coordinates of the point clouds within the same voxel are calculated to obtain the initial features ${{F}_{v}}\in {{\mathbb{R}}^{{{N}_{v}}\times 3}}$. Similar to the pillar branch, a four-stage 3D encoder is employed in the first half of the voxel branch. At the end of each stage, focal convolution \cite{36} is introduced, which adaptively changes kernel shapes during training to retain valuable foreground information, forming focal sparse convolution blocks (FSB). To ensure consistent resolution and similar levels of abstraction in the horizontal direction for the intermediate layer features extracted by the dual branches, the corresponding SCB and FSB maintain the same horizontal convolution kernel size, stride, padding, and output feature dimensions. This facilitates the subsequent element-wise alignment and sparse fusion of features. Ultimately, 8$\times $downsampled sparse voxels $F_{v}^{4}\in {{\mathbb{R}}^{N_{v}^{4}\times C_{v}^{4}}}$ are obtained, where $C_v^4$ represents the number of feature channels in the fourth layer of the Voxel branch.

\begin{figure}[h]
    \centering
    \includegraphics[width=0.45\textwidth,height=0.3\textwidth]{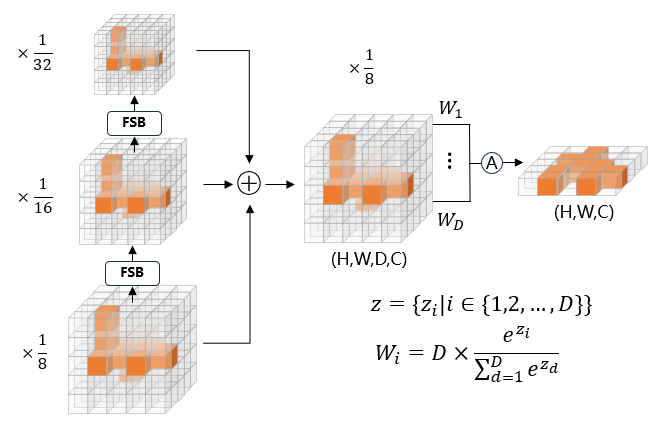}
    \caption{The architecture of multi-scale feature aggregation module}
    \end{figure} 

To better overcome the spatial structural similarity between objects and backgrounds, and the complex distribution of objects in the 3D space of the airborne scene, we design a multi-scale feature aggregation module for the second half of the branch. This module provides multi-scale 3D receptive fields through additional downsampling layers and efficiently integrates vertical features using weighted elevation compression, as illustrated in Figure 4. First, $F_{v}^{4}$ is sequentially processed through two FSBs to obtain 16$\times $and 32$\times $ downsampled sparse voxels $F_{v}^{5}\in {{\mathbb{R}}^{N_{v}^{5}\times C_{v}^{4}}}$ and $F_{v}^{6}\in {{\mathbb{R}}^{N_{v}^{6}\times C_{v}^{4}}}$ along with corresponding indices $I_{v}^{5}\in {{\mathbb{N}}^{N_{v}^{5}\times 3}}$ and $I_{v}^{6}\in {{\mathbb{N}}^{N_{v}^{6}\times 3}}$. Compared to $F_{v}^{4}$, they have a wider receptive field and lower spatial resolution. The above features are then element-wise added based on respective indices and sampling ratios, resulting in sparse voxels $F_{v}^{\text{ }\!\!'\!\!\text{ }}\in {{\mathbb{R}}^{N_{v}^{\text{ }\!\!'\!\!\text{ }}\times C_{v}^{\text{ }\!\!'\!\!\text{ }}}}$ and indices $I_{v}^{\text{ }\!\!'\!\!\text{ }}\in {{\mathbb{N}}^{N_{v}^{\text{ }\!\!'\!\!\text{ }}\times 3}}$ with multi-scale receptive fields.

Current Voxel-based 3D object detection networks typically compress output features of the encoder by concatenating channels along the vertical direction. However, in airborne scenes, the vertical dimension of the detection area is extensive, and such concatenation increases computational resource consumption. Furthermore, the distribution of objects of interest is regional, and the amount of useful information varies significantly across different elevation levels. 

To address these issues, we introduce a learnable parameter $z\in {{\mathbb{R}}^{D\times 1}}$, where $D$ represents the number of dimensions in the vertical direction of the voxel features. The sparse weights $W\in {{\mathbb{R}}^{D\times 1}}$ for different heights are obtained through Softmax normalization of the learnable parameter $z$, with the meaning of importance weights of voxels at different vertical heights for detection scene:

\begin{equation}
W=\{{{W}_{i}}=D\times {{e}^{{{z}_{i}}}}/\mathop{\sum }_{j=1}^{D}{{e}^{{{z}_{j}}}}|\forall i\in \{1,2...D\}\}
\end{equation}
 
Weighted sparse voxel features $F_{v}^{b}\in {{\mathbb{R}}^{{{N}_{b}}\times C_{v}^{4}}}$ and indices $I_{v}^{b}\in {{\mathbb{R}}^{{{N}_{b}}\times 2}}$ are then obtained through summation and compression along the vertical dimension as
\begin{equation}
F_{v}^{b}=\{\sum\limits_{i\in {{F}_{xy}}}{F_{v}^{\text{ }\!\!'\!\!\text{ }}[i,:]\cdot W[I_{v}^{\text{ }\!\!'\!\!\text{ }}[i,3]]}|(x,y)\in I_{v}^{b}\}
\end{equation}
\begin{equation}
I_{v}^{b}=\{({{x}_{i}},{{y}_{i}})|i\in \text{Unique}(I_{v}^{\text{ }\!\!'\!\!\text{ }}[:,(1,2)])\}
\end{equation}
where 
\begin{equation}
{{F}_{xy}}=\{i|I_{v}^{\text{ }\!\!'\!\!\text{ }}[i,1]=x,I_{v}^{\text{ }\!\!'\!\!\text{ }}[i,2]=y,\forall i\in \{1...N_{v}^{\text{ }\!\!'\!\!\text{ }}\}\}
\end{equation}	

represents the set of all vertical voxel features at the same horizontal coordinates $(x,y)$  in the BEV, Unique represents the de-duplication operator, and $N_b$ represents the number of unique BEV positions retained after height compression, which is the total number of deduplicated horizontal coordinates $(x,y)$. Finally, $F_{v}^{b}$ is transformed into dense voxel features $F_{v}^{D}\in {{\mathbb{R}}^{H\times W\times C}}$.

\subsection{Multi-Level Feature Fusion Mechanism}
The complex-heterogeneous characteristics of airborne HPCs can be effectively encoded by the pillar-voxel dual-branch structure, which emphasizes different aspects of the data. However, the key to enhancing model performance lies in fully leveraging these complementary features and achieving organic fusion of the dual-branch features. To address this, we design a multi-level feature fusion mechanism that adaptively integrates features from different branches, ensuring the full exploitation of interrelated data during the fusion process. This mechanism comprises two unique modules: a sparse feature fusion module, which fuses the intermediate features extracted by the FSB and SCB, and a patch-wise feature fusion module, which fuses the output features of the dual branches.

Compared to a vanilla dual-branch encoder without feature exchange during extraction, sparse feature fusion in intermediate layers establishes information exchange between branches, enabling the network to dynamically adjust feature correlations. Furthermore, the spectral and vertical structure information extracted by the pillar branch enriches the intermediate voxel features, which provides spectral confidence criterion and a more effective classification basis for the subsequent refinement network. 

Specifically, the sparse feature fusion module uses the pillar branch as an auxiliary, adding element-wise intermediate layer features to the voxel branch. Since the feature maps of both branches at the same stage maintain the same spatial structure and feature dimensions in the horizontal direction, they can be aligned in the BEV based on their respective position indices ${{I}_{p}}$ and ${{I}_{v}}$, calculating the associated indices set ${{I}_{p2v}}$:
\begin{equation}
{{I}_{p}}=\{{{I}_{{{p}_{i}}}}=\left( {{h}_{i}},{{w}_{i}} \right)\in {{\mathbb{N}}^{2}}|i\in \left\{ 1,\ldots ,{{N}_{p}} \right\}\}
\end{equation}

\begin{equation}
{{I}_{v}}=\{{{I}_{{{v}_{i}}}}=\left( {{x}_{i}},{{y}_{i}},{{z}_{i}} \right)\in {{\mathbb{N}}^{3}}|i\in \left\{ 1,\ldots ,{{N}_{v}} \right\}\}
\end{equation}

\begin{equation}
{{I}_{p2v}}=\left\{ \left( k,l \right)\text{ }\!\!|\!\!\text{ }k\in \left\{ 1,\ldots ,{{N}_{v}} \right\},l\in \left\{ 1,\ldots ,{{N}_{p}} \right\},\left( {{x}_{k}},{{y}_{k}} \right)=\left( {{h}_{l}},{{w}_{l}} \right) \right\}
\end{equation}

The ${{I}_{{{p}_{i}}}}$and${{I}_{{{v}_{i}}}}$ represents the indices of the i th sparse pillar and voxel, respectively. According to the ${{I}_{p2v}}$, an association matrix ${{M}_{kl}}\in {{\mathbb{N}}^{{{N}_{v}}\times {{N}_{p}}}}$ can be constructed to sum the pillar and voxel features at corresponding positions as follows:
\begin{equation}
{{M}_{kl}}=\left\{ \begin{matrix}
   1\text{ }\left( k,l \right)\in {{I}_{p2v}}  \\
   0\text{ otherwise}  \\
\end{matrix} \right.
\end{equation}

\begin{equation}
{{F}_{v}}={{F}_{v}}+{{M}_{kl}}\centerdot {{F}_{p}}
\end{equation}

The output feature maps $F_{v}^{D}$ and $F_{p}^{D}$ from both branches are spatially aligned due to the same spatial structure of the initial input data and similar encoder designs. However, differences in multi-sensor imaging modalities cause unavoidable alignment errors, resulting in geometric distortions (spectral shifts) of the features. Additionally, learning the correlations between heterogeneous features of the two branches and selecting important feature channels are crucial challenges for enhancing the algorithm's data representation ability. Simple concatenation or summation fusion methods struggle to overcome these issues, hence a patch-wise feature fusion module was designed as shown in Figure 5.

The primary advantage of this module lies in its ability to achieve precise feature alignment and channel selection based on learned modal associations, adaptively highlighting more valuable features in the neighborhood. Specifically, fully connected layers first encode BEV features as query (Q), and the corresponding $k\times k$ size (default $k$=3) neighborhood pillar features are encoded as key (K) and value (V)with added positional encoding. Through multi-head Q and V outer products and Softmax operations, the attention weight matrix between them is calculated. Then, this matrix is multiplied by value and concatenated with voxel features after passing through fully connected layers, yielding the aligned BEV feature map ${{F}_{align}}\in {{\mathbb{R}}^{H\times W\times 2C}}$.

The calculation process for channel selection is similar to feature alignment.  The attention matrix is computed across different feature channels of the $k\times k$ image patch, with K, Q, and V derived from the same source, implementing a self-attention algorithm on the channel dimension. The final output is the fused BEV feature map ${{F}_{bev}}\in {{\mathbb{R}}^{H\times W\times 2C}}$.

The calculation process of the patch-wise feature fusion module is shown in Equation 11, where FC denotes the fully connected layer, ${{{\rm{F}}_{n \times n}}}$ represents the sliding window operation with a window size of $n$), and Concat indicates the concatenation operation.

\begin{equation}
\left\{ \begin{array}{l}
{{\rm{F}}_{1 \times 1}}{\rm{(}}{F_{align}}{\rm{)}} = {\rm{Concat}}({\rm{Softmax(}}\frac{{{\rm{FC[}}{{\rm{F}}_{1 \times 1}}{\rm{(}}F_v^D{\rm{)]FC[}}{{\rm{F}}_{k \times k}}{\rm{(}}F_p^D{\rm{)}}{{\rm{]}}^T}}}{{\sqrt {\mathrm{d}} }}{\rm{)FC[}}{{\rm{F}}_{k \times k}}{\rm{(}}F_p^D{\rm{)]}},{{\rm{F}}_{1 \times 1}}{\rm{(}}F_v^D{\rm{)}})\\
{{\rm{F}}_{1 \times 1}}{\rm{(}}{F_{bev}}{\rm{)}} = {\rm{Softmax(}}\frac{{{\rm{FC[}}{{\rm{F}}_{k \times k}}{\rm{(}}{F_{align}}{\rm{)]FC[}}{{\rm{F}}_{k \times k}}{\rm{(}}{F_{align}}{\rm{)}}{{\rm{]}}^T}}}{{\sqrt {\mathrm{d}} }}{\rm{)FC[}}{{\rm{F}}_{k \times k}}{\rm{(}}{F_{align}}{\rm{)]}}
\end{array} \right.
\end{equation}


\begin{figure}[h]

    \centering
    \includegraphics[width=1\textwidth]{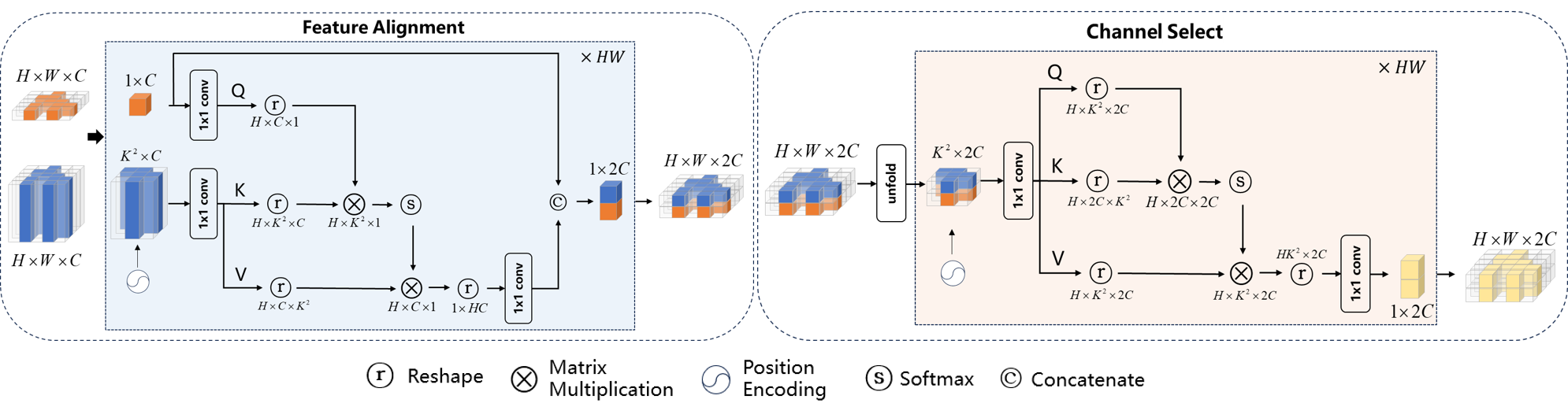}
    \caption{The architecture of patch-wise feature fusion module }
    \end{figure} 
    
\section{Experiments and analysis}
\subsection{Datasets}

\textbf{1) HIT Campus Dataset: }The dataset contains the following two sets of data: scenes containing vehicle and box objects were collected on September 27, 2022, using a UAV equipped with a Headwall-Nano hyperspectral camera and a VLP-16 lidar; scenes containing vehicle and tent objects were collected on July 7, 2023, using UAVs equipped with the same hyperspectral camera and a RIEGL-MiniVUX lidar. The data covers the Harbin Institute of Technology Science Park and its surrounding areas, captured at an altitude of 80m. The total number of point clouds is 5,349,316 and 8,797,764, with point cloud densities of 84 and 169, respectively. The HSI has a spatial resolution of 0.8 meters and includes 273 bands covering the spectral range of 400-1000nm. The HPCs was obtained by registering HSI with the digital surface model (DSM) generated from the point clouds using the approximate nearest neighbor (ANN) algorithm. Due to hardware limitations, the data needed to be split into multiple small scenes for network training. The data was split based on different flight paths and rotated at various angles. A sliding window method was employed with a step size of 15 meters and a window size of ${\text{30m}\times \text{30m}}$ along the BEV direction, resulting in 5421 training and 2654 test scenes. To validate the algorithm's robustness and increase task difficulty, we added some fake objects as interference in the scene. Table 1 lists the number of training and testing samples for each class. 
\begin{table}[!ht]
\centering
\footnotesize
\caption{ SAMPLE SIZE OF HIT CAMPUS DATASET TRAINING AND TESTING DATASETS}
    \begin{tabular}{ccccc}
    \hline
        No. & Class Name & Training & Testing & Color \\ \hline
        1 & Vehicle & 14794 & 6289 & \includegraphics[width=0.6cm,height=0.18cm]{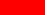} \\ 
        2 & Tent & 717 & 433 & \includegraphics[width=0.6cm,height=0.18cm]{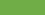} \\ 
        3 & Box & 1104 & 721 & \includegraphics[width=0.6cm,height=0.18cm]{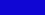} \\ \hline
        ~ & Total & 16615 & 8578 & ~ \\ \hline
    \end{tabular}
\end{table}

\begin{figure}

    \centering
    \includegraphics[width=0.75\textwidth]{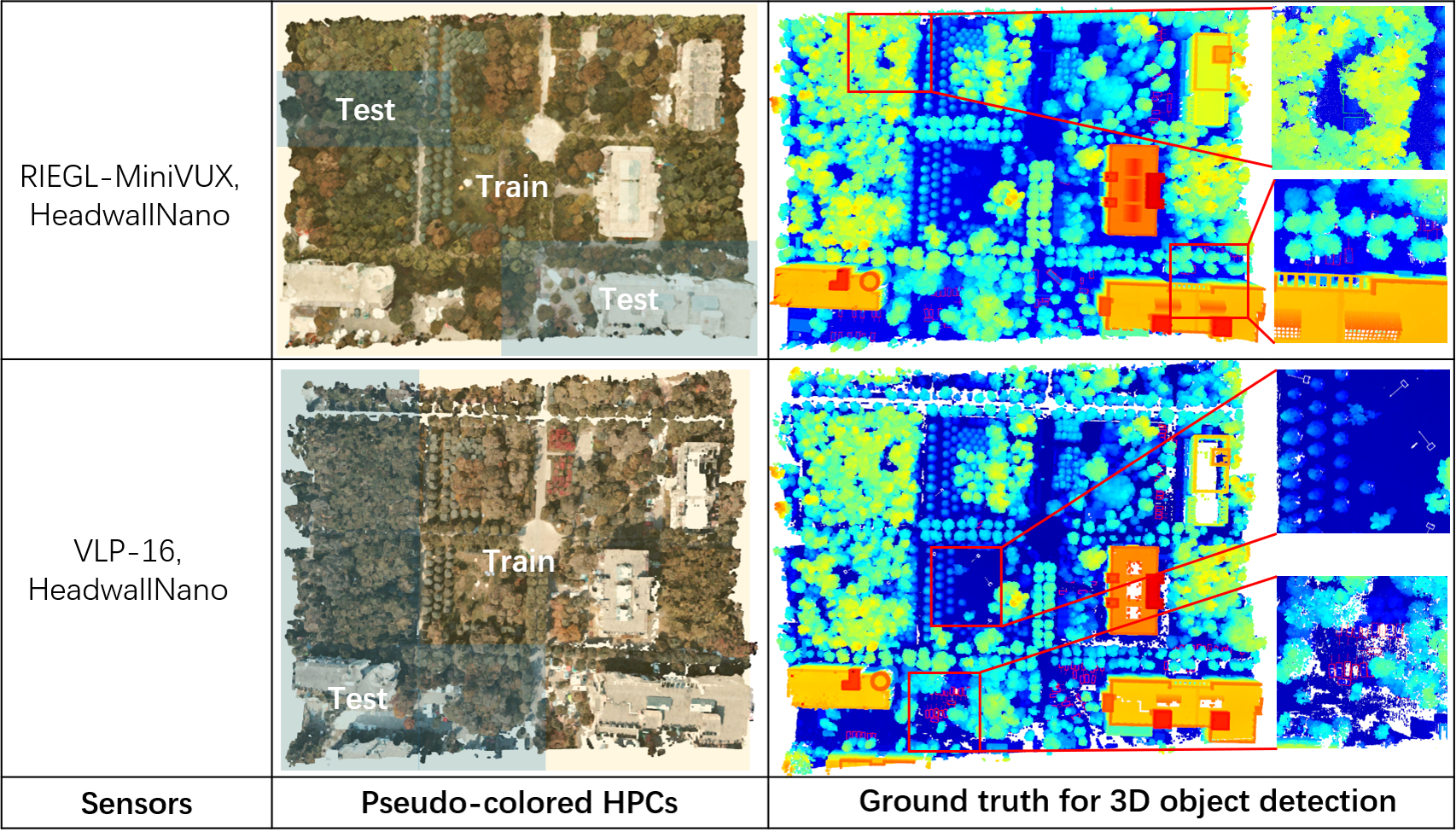}
    
    \caption{Visualization of HIT Campus dataset, including pseudo-colored HPCs and ground truth for 3D object detection, where training regions are shown in orange and test regions in blue}
\end{figure}

\textbf{2) Houston Dataset: } The original data for this dataset was provided by the 2018 IEEE GRSS Data Fusion Challenge, collected on February 16, 2017. The HSI has a spatial resolution of 1 meter and contains 48 bands covering the spectral range of 380-1050 nm. We selected and labeled the scene numbered '272056\_3289689', which contains 13,955,886 point clouds with a density of 39$\text{pts/}{\text{m}}^{2}$. The scene's objects of interest are vehicles. The data preprocessing process is the same as that of the HIT Campus Dataset, resulting in 3469 training scenes and 1129 test scenes. Figure 7 shows the pseudo-color point cloud, 3D object detection labels, and the division of training and testing areas of this dataset. For the same purpose as the HIT Campus Dataset, we converted some scene objects into fake ones. The training scenes contain 20320 vehicle objects, while the test scenes contain 3065 vehicle objects.

\textbf{3) HIT-L2 Dataset: } We collected point cloud and image data of Harbin Institute of Technology campus on June 26, 2024, using a drone equipped with Zenmuse L2 LiDAR and RedEdge-MX multispectral camera. The scenes contain three types of objects: vehicles, canopies, and tents. Multi-view multispectral images underwent spectral super-resolution processing to generate pseudo point clouds\cite{61}, which were registered with LiDAR point clouds to assign spectral information. This process yielded hyperspectral point cloud data containing 40 spectral bands covering wavelengths from 398-1000nm. Following the same construction approach as the HIT-Campus dataset, we generated 941 training scenes and 1408 test scenes.

\subsection{Experimental Setup}

\textbf{1) Evaluation Metrics: }In our experiments, we followed the calculation methods proposed by the public Kitti dataset \cite{51} to evaluate the accuracy of 3D object detection algorithms using three metrics: Average Precision for BEV ($\text{A}{{\text{P}}_{\text{BEV}}}$), Average Precision for 3D ($\text{A}{{\text{P}}_{\text{3D}}}$), and Average Orientation Similarity (AOS). The AP metrics combine precision and recall to comprehensively reflect the network's detection capabilities. The above three metrics are calculated from the perspectives of BEV, 3D space, and angle, respectively. For direction-insensitive categories such as tents and boxes, their AOS direction similarity metrics were calculated using equation (12) to alleviate direction penalties. We set 'easy', 'moderate', and 'hard' difficulty levels based on the number of point clouds in different object categories, with the distribution of objects in various difficulty levels shown in Figure 8. To evaluate the algorithm's detection capability for occluded objects, we divided the objects into 'occlusion-free', 'part occlusion', and 'high occlusion' levels based on the object neighborhood's occlusion density. The distribution and corresponding object quantities under different occlusion levels for each dataset are listed in TABLE 2.

\begin{equation}
s(r)=\frac{1}{\left| D(r) \right|}\sum\limits_{i\in D(r)}{\frac{1+\cos (4*\Delta \theta )}{2}}{{\delta }_{i}}
\end{equation}

\begin{table}[!ht]
\centering
\footnotesize
\caption{ CLASSIFICATION OF OCCLUSION LEVELS}
\label{tab1}
    \begin{tabular}{c|ccc|ccc}
    \hline
        DATASET & \multicolumn{3}{c|}{HIT Campus} & \multicolumn{3}{c}{Houston2018} \\ \hline
        Canopy range & 0 & (0,0.5] & (0.5,1] & 0 & (0,0.3] & (0.3,1] \\ \hline
        Object Num. & 7766 & 8187 & 6116 & 32373 & 3048 & 2980 \\ \hline
    \end{tabular}
\end{table}

\begin{figure}

    \centering
    \includegraphics[width=0.75\textwidth]{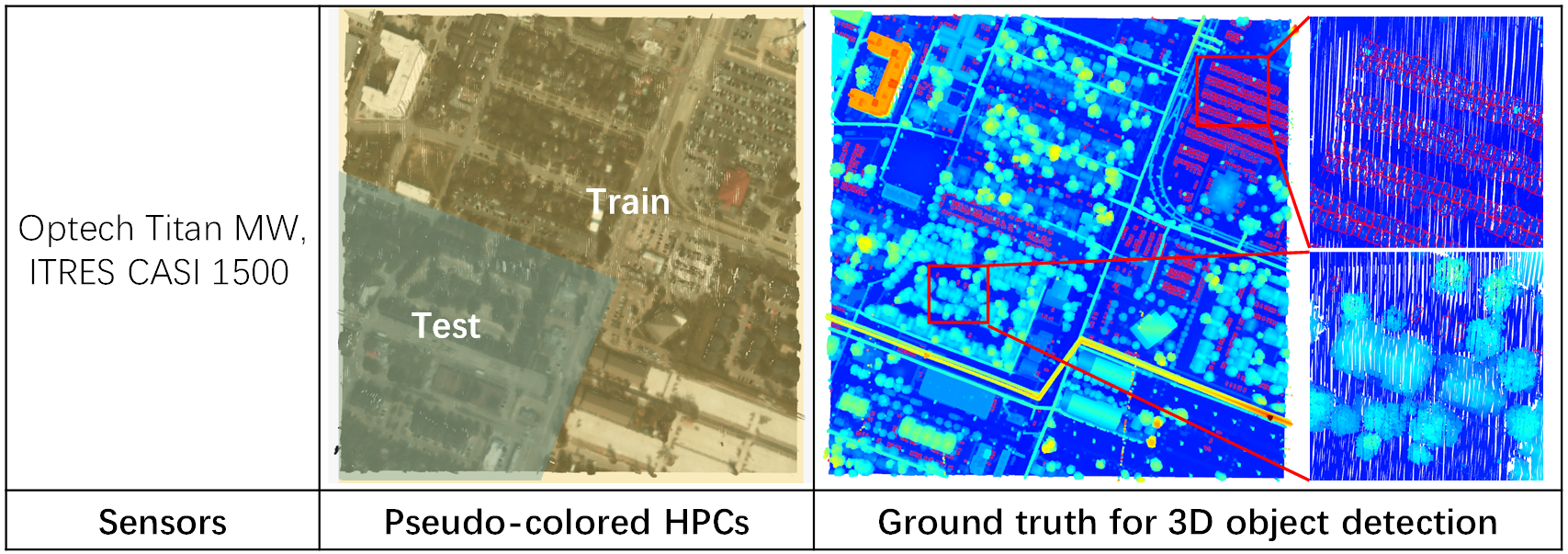}
    \caption{Visualization of Houston 2018 dataset, including pseudo-colored HPCs and ground truth for 3D object detection, where training regions are shown in orange and test regions in blue}
\end{figure}

\textbf{2)Implementation Details: }Our algorithm was implemented using the OpenPCDet\cite{62} codebase based on the PyTorch framework. For both datasets, we employed the Distributed Data Parallel strategy (4×TitanXP GPUs, batch size of 2 per GPU), AdamW optimizer (learning rate 0.005, weight decay 0.01), and OneCycle scheduling strategy (peak learning rate 0.05, momentum range [0.85, 0.95]), training for a total of 30 epochs. The loss functions are structured in two network stages: the region proposal network employs Focal classification loss, Smooth L1 regression loss, and direction classification loss, while the refinement network utilizes cross-entropy classification loss, Smooth L1 regression loss, and 3D corner regression loss. All hyperparameters for these loss components follow the OpenPCdet settings. The input voxel size was set to (0.1m, 0.1m, 0.1m), the input pillar size was set to (0.1m, 0.1m, 16m), and the detection range was configured as (-25.6~25.6m, -25.6~25.6m, 0~16m).

\begin{figure}[!ht]
    \centering
    \includegraphics[width=0.45\textwidth,height=0.3\textwidth]{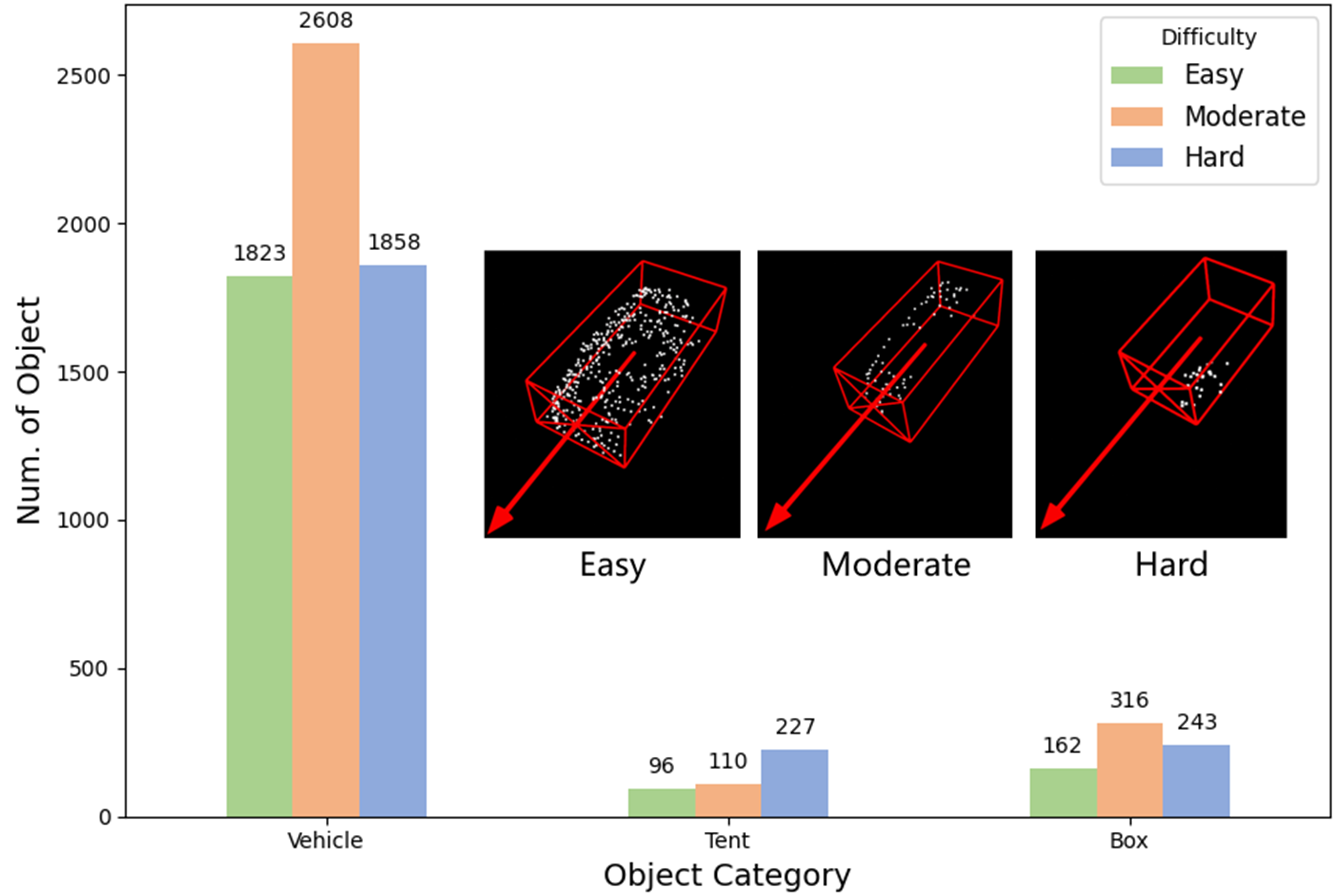}
    \caption{Distribution of different object categories by difficulty}
\end{figure} 

\subsection{Comparative Analysis with the State-of-the-Art}
To demonstrate the effectiveness of the PiV-AHPC, several representative algorithms from the point cloud object detection domain were chosen for quantitative and qualitative comparisons on the HIT Campus Dataset and the Houston2018 Dataset. These include PillarNet \cite{40}, which is pillar-based; PV-RCNN++ \cite{38}, which is point-voxel based; Voxel-RCNN \cite{35}, which is voxel-based; CenterPoint \cite{43}, which utilizes center predictions; PartA2\cite{52}, which is basedd on a refinement network; TransFusion-L \cite{53}, which is based on a transformer architecture; and VoxelNext \cite{48}, which focuses on sparse prediction.

\textbf{1) Performance on HIT Campus test set:} Table 3 presents the detection accuracy of the proposed method compared to state-of-the-art (SOTA) algorithms. It can be observed that PiV-AHPC demonstrates outstanding detection performance. Compared to all other methods, its mAP scores across different sample difficulties (especially in moderate and hard samples) show significant improvement. Specifically, compared to the best results of other methods, PiV-AHPC achieves improvements in detection precision: 5.33\%, 16.71\%, and 7.58\% for BEV, 3D, and AOS metrics, respectively, at the easy level; 12.22\%, 16.92\%, and 14.41\% at the moderate level; and 14.98\%, 14.09\%, and 16.21\% at the hard level. This indicates that the algorithm can effectively distinguish and detect the three types of objects from complex and noisy backgrounds. Comparing the performance of all methods on each object category reveals that PiV-AHPC achieves optimal detection performance on tents and boxes. This is attributed to its unique dual-branch encoding architecture and multi-level feature fusion mechanism, which better capture geometric and spectral characteristics of objects. For vehicle detection, our method shows competitive performance with the baseline method, generally outperforming other comparison methods but not consistently surpassing the baseline across all metrics. This indicates that while PiV-AHPC demonstrates strong overall performance, there remains room for improvement in detecting specific object categories.

Figure 9 and Figure 10 visualize the detection results of the methods on the large-scale scene. These results indicate that the proposed method can accurately classify and locate the objects of interest, demonstrating strong anti-interference capabilities with fewer occurrences of false positives and false negatives.

\textbf{2) Performance on Houston2018 test set:} TABLE 4 presents the performance of PiV-AHPC compared to the SOTA algorithms on the Houston2018 test set. The proposed method achieves significant improvements across all difficulty levels: easy, moderate, and hard, as well as IoU thresholds ranging from 0.5 to 0.95, demonstrating superior classification and localization capabilities. Specifically, compared to the best results achieved by other algorithms, PiV-HCCD shows improvements of 5.2\%, 13.15\%, 10.48\%, and 5.51\% in $\text{A}{{\text{P}}_{\text{3D}}}$, respectively. The algorithm demonstrates outstanding performance on moderate and hard objects, showcasing a powerful understanding capability for complex scenes. Notably, PiV-AHPC and several comparison methods achieve better performance on moderate objects than on easy ones. This may be due to the dataset distribution, where moderate objects in test set exhibit more representative features, allowing models to adapt more effectively. Figure 11 illustrates the visualization of the detection results of all methods in large-scale scenes. We can intuitively observe that the proposed method effectively detects most objects while minimizing false positive, demonstrating PiV-AHPC's strong performance and robustness in large-scale scenes.

\begin{figure}
    \centering
    \subfigure[Ground Truth]{
        \includegraphics[width=0.18\textwidth]{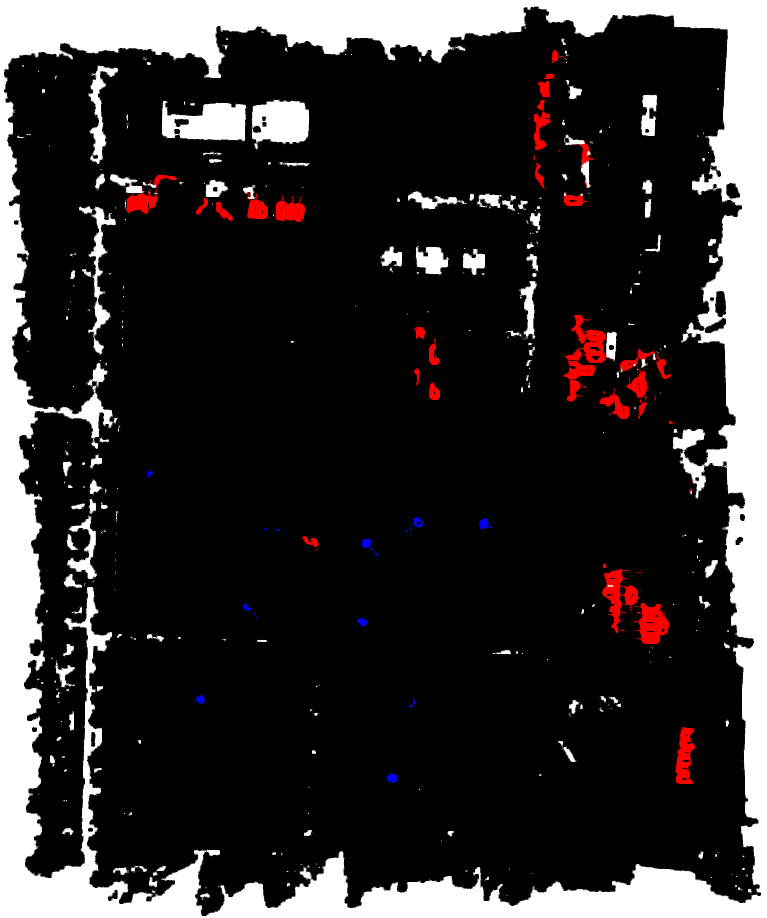}
    }%
    \subfigure[PartA2]{
        \includegraphics[width=0.18\textwidth]{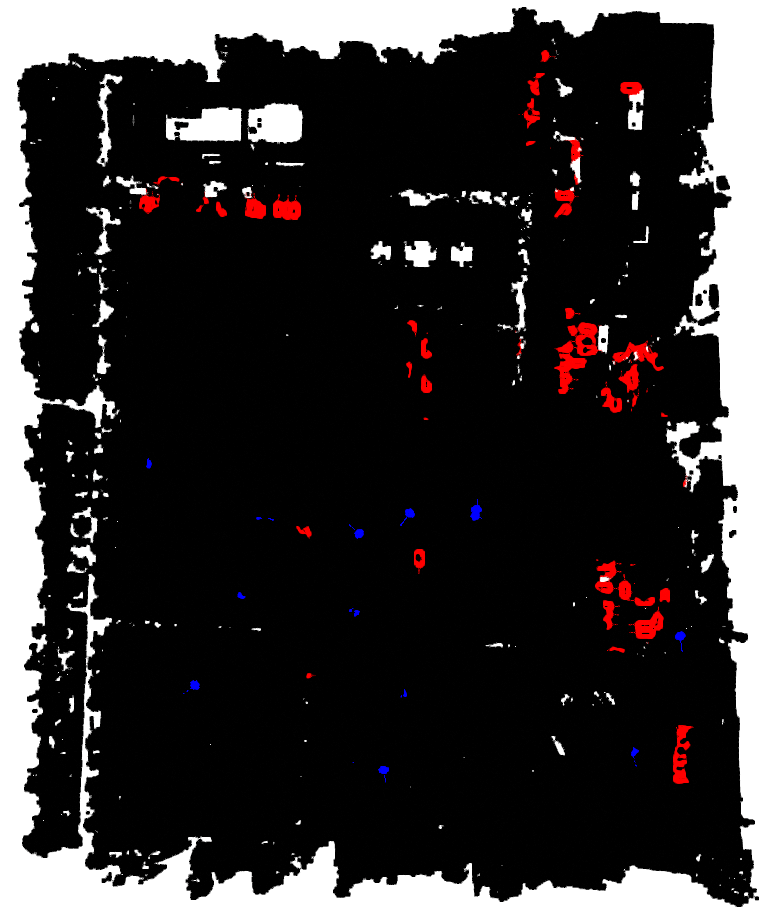}
    }%
    \subfigure[CenterPoint]{
        \includegraphics[width=0.18\textwidth]{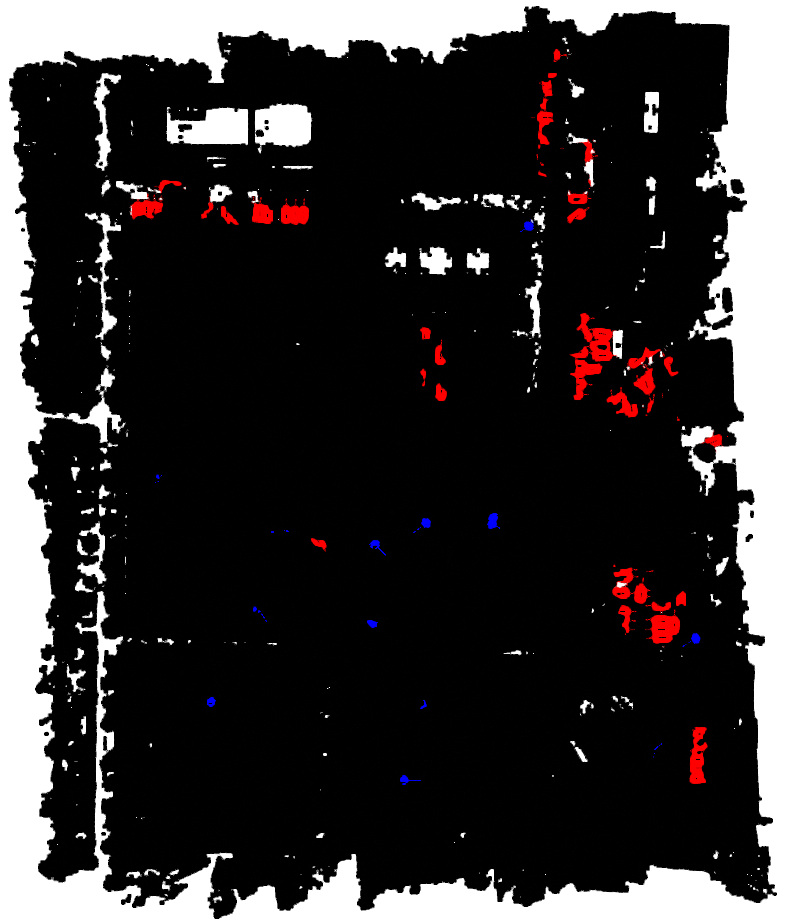}
    }%
    \subfigure[PillarNet]{
        \includegraphics[width=0.18\textwidth]{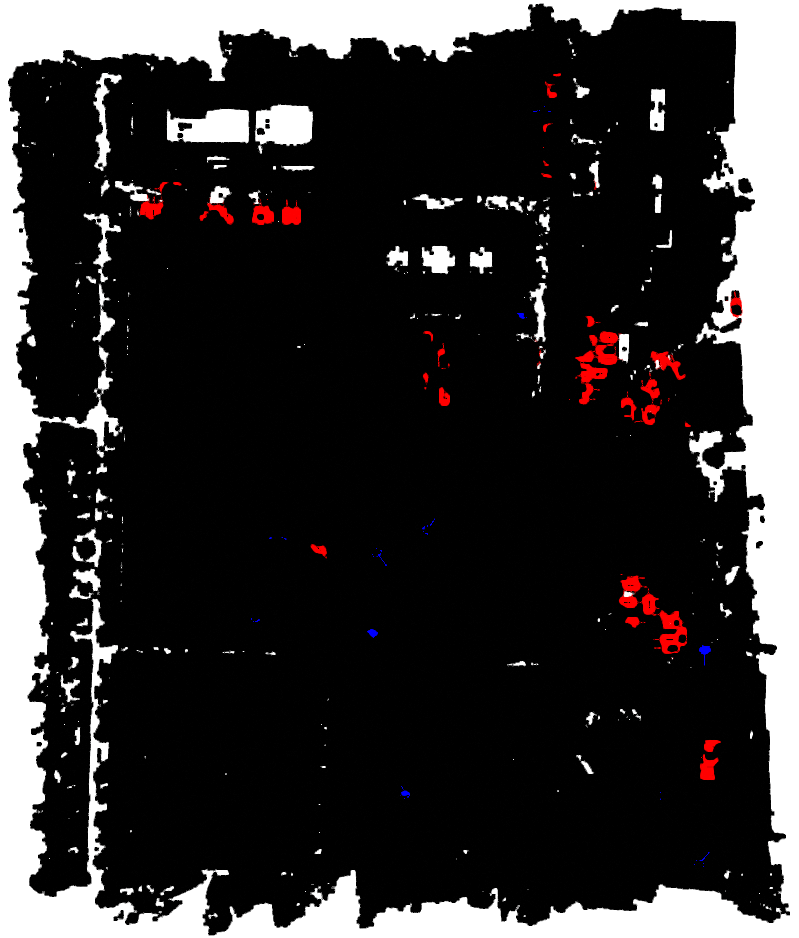}
    }%
    
    \subfigure[PV-RCNN++]{
        \includegraphics[width=0.18\textwidth]{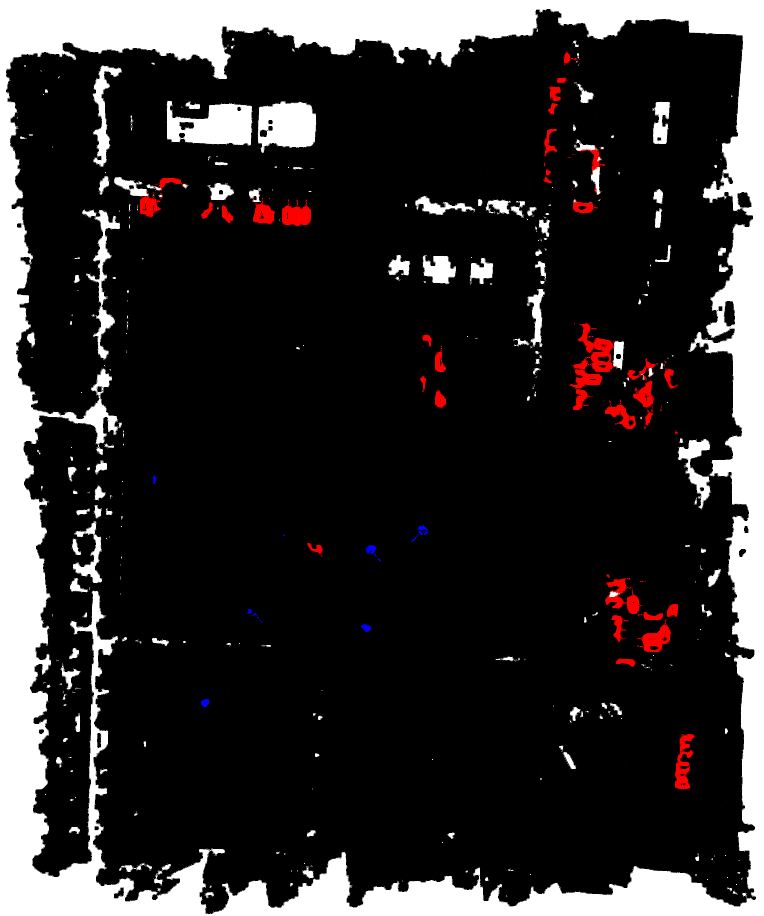}
    }%
    \subfigure[Voxel-RCNN]{
        \includegraphics[width=0.18\textwidth]{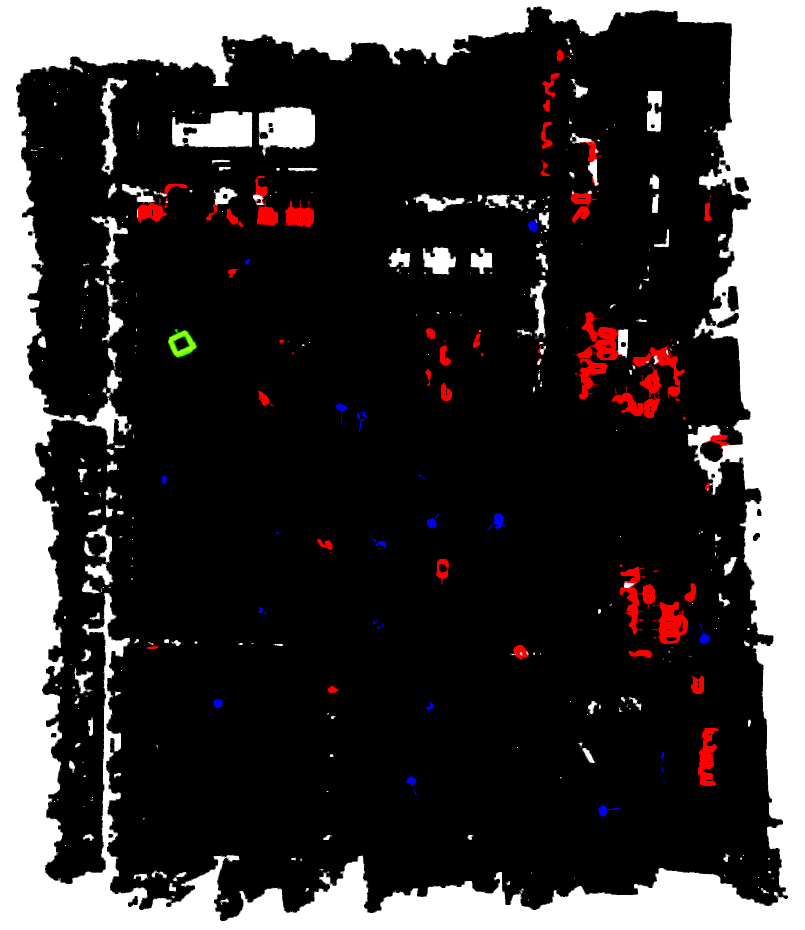}
    }%
    \subfigure[TransFusion-L]{
        \includegraphics[width=0.18\textwidth]{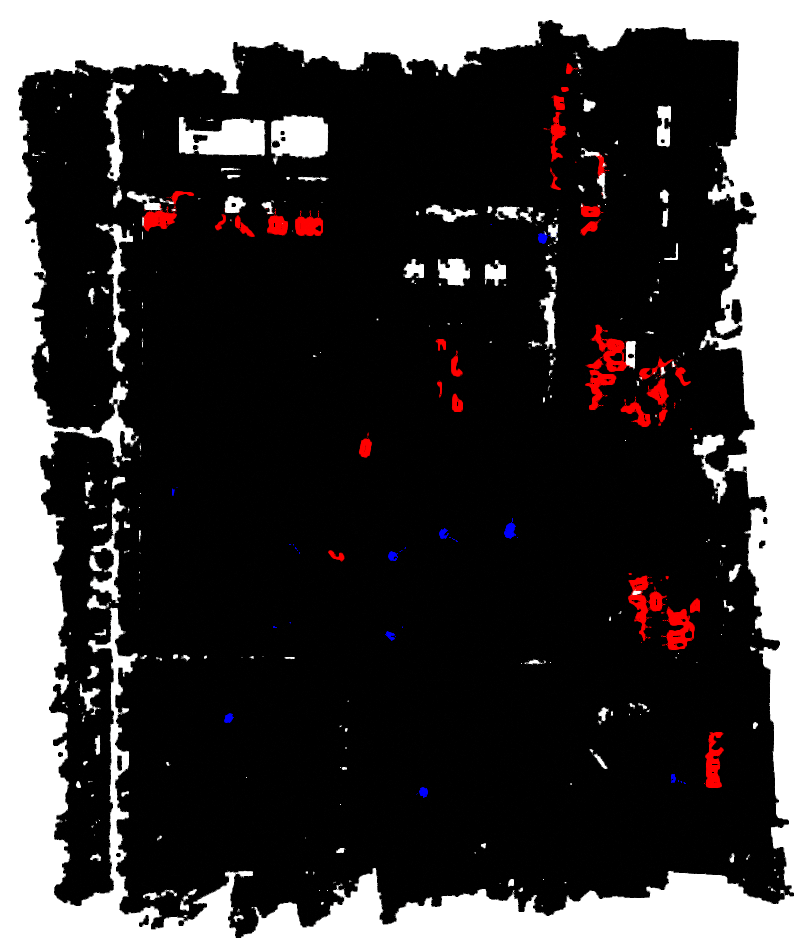}
    }%
    \subfigure[VoxelNext]{
        \includegraphics[width=0.18\textwidth]{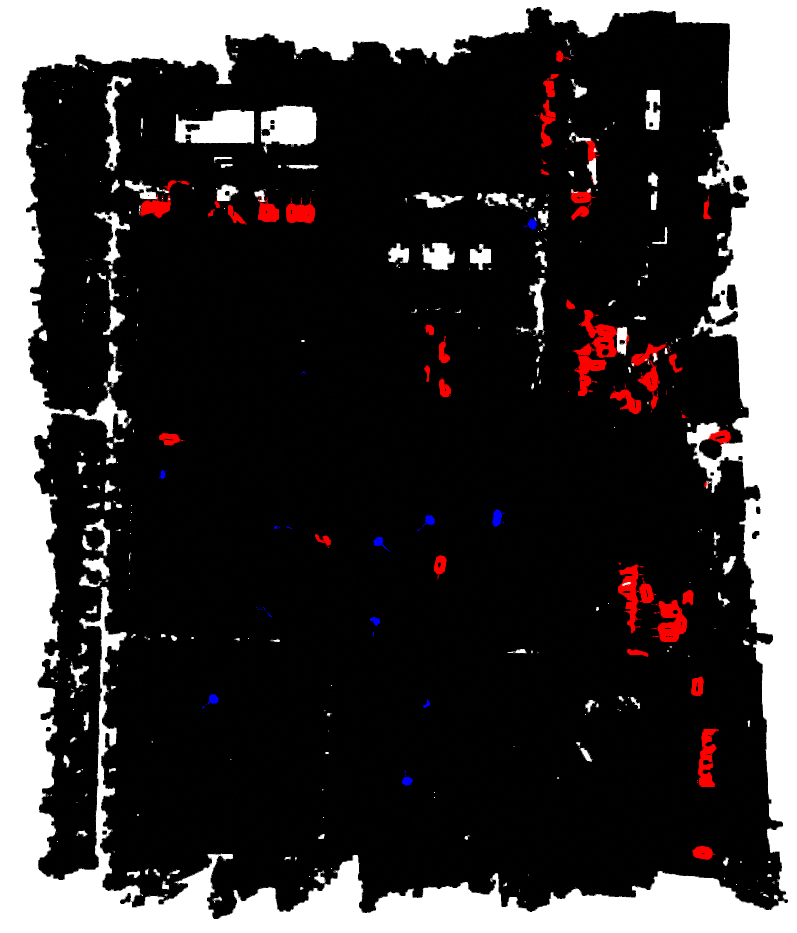}
    }%
    \subfigure[Proposed]{
        \includegraphics[width=0.18\textwidth]{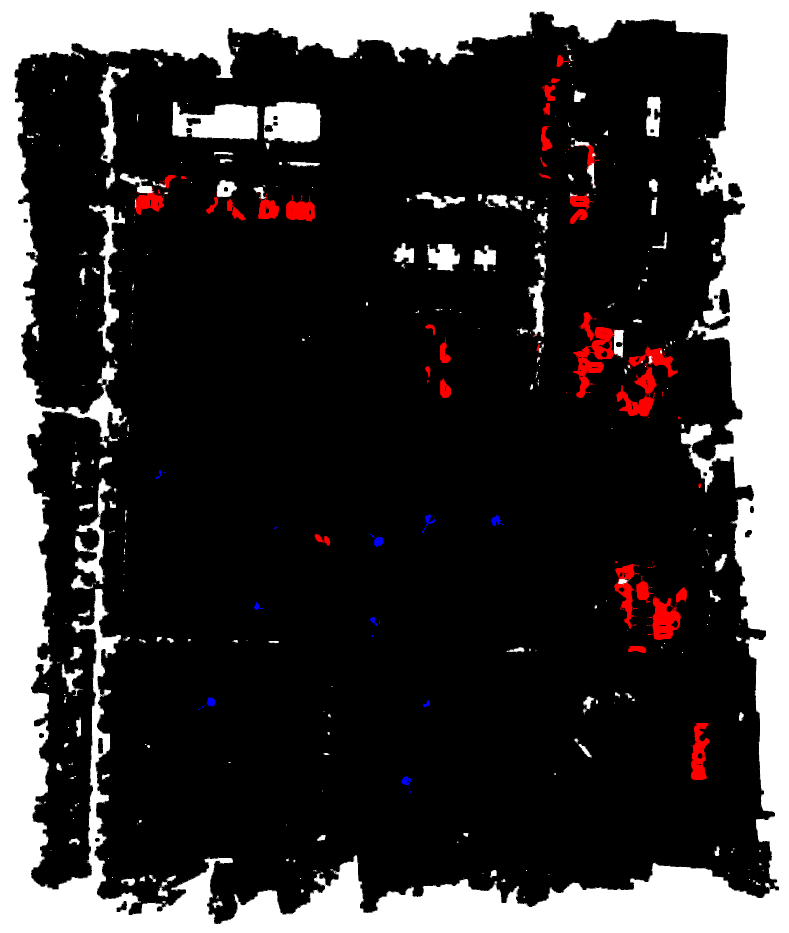}
    }%
    \caption{Visualization of the predicted 3D bounding boxes in HIT Campus dataset scene 1, where red represents vehicles, green represents tents, and blue represents boxes}
\end{figure}

\begin{figure}
    \centering
    \subfigure[Ground Truth]{
        \includegraphics[width=0.18\textwidth]{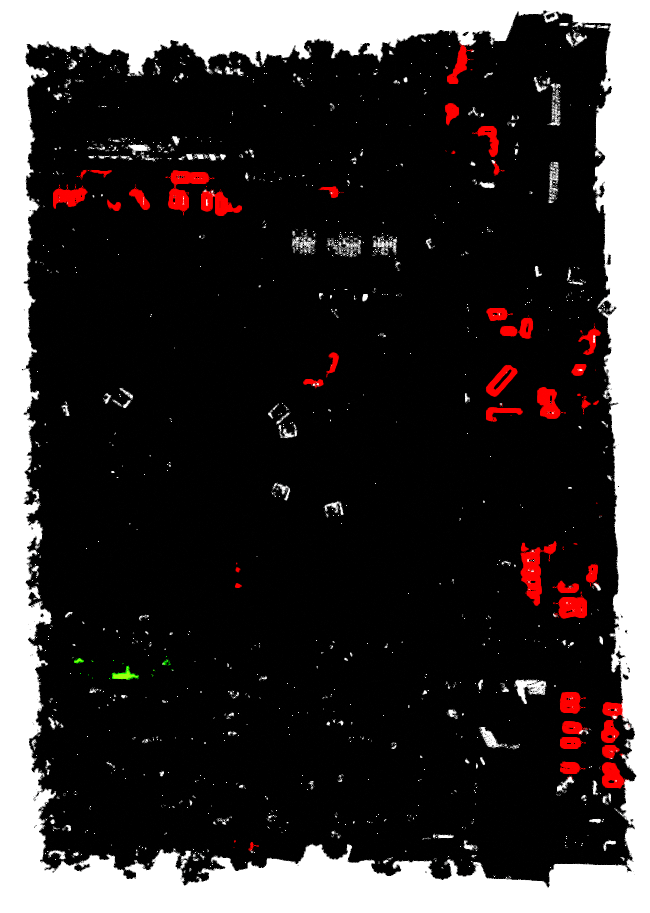}
    }%
    \subfigure[PartA2]{
        \includegraphics[width=0.18\textwidth]{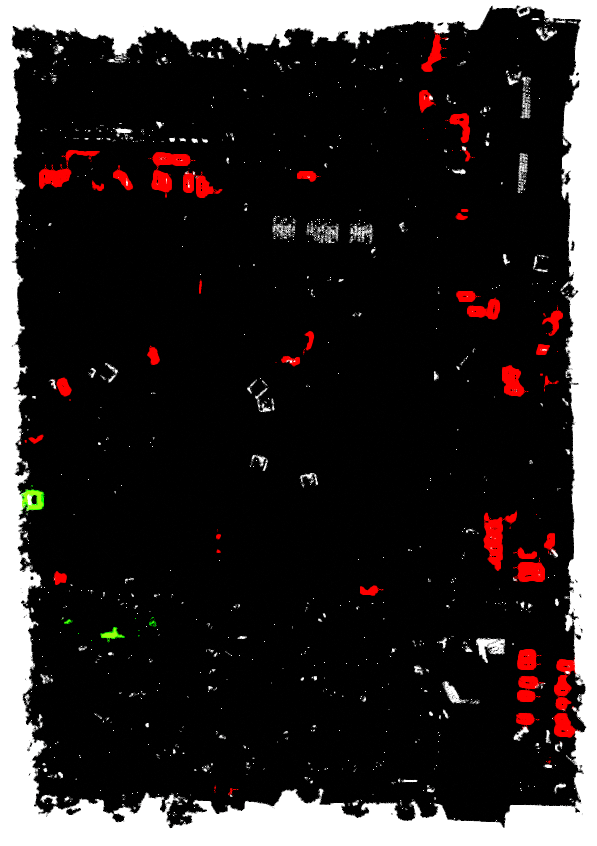}
    }%
    \subfigure[CenterPoint]{
        \includegraphics[width=0.18\textwidth]{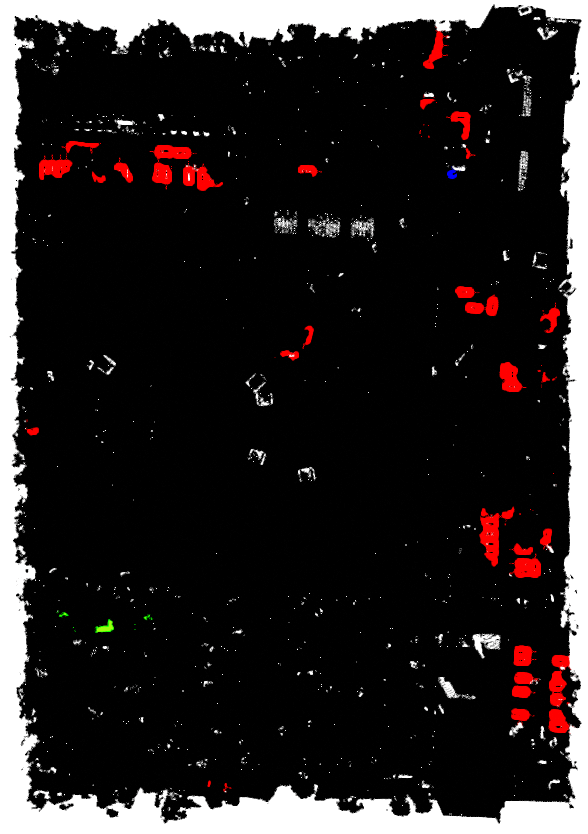}
    }%
    \subfigure[PillarNet]{
        \includegraphics[width=0.18\textwidth]{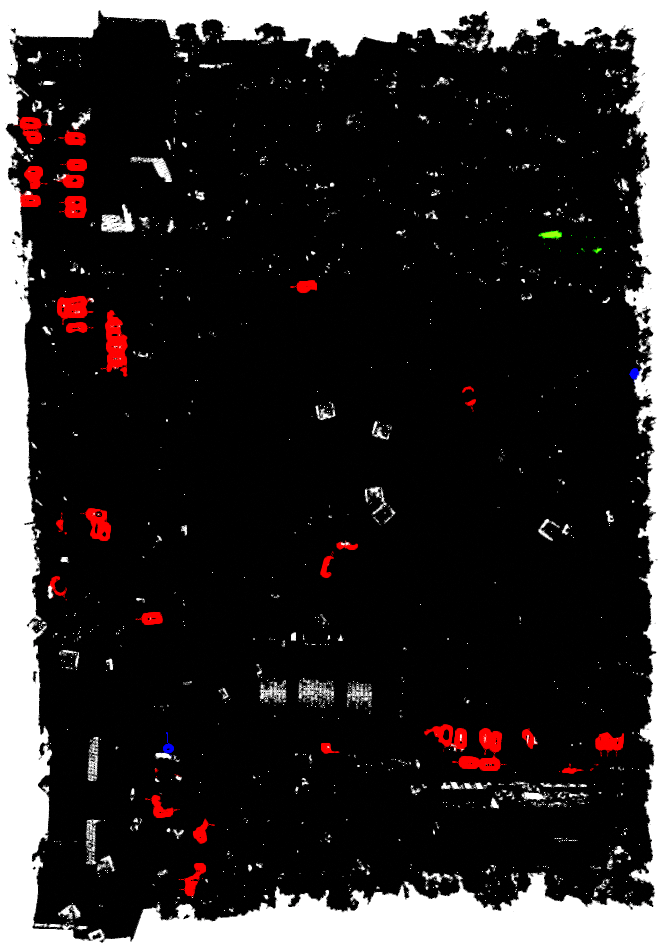}
    }%
    
    \subfigure[PV-RCNN++]{
        \includegraphics[width=0.18\textwidth]{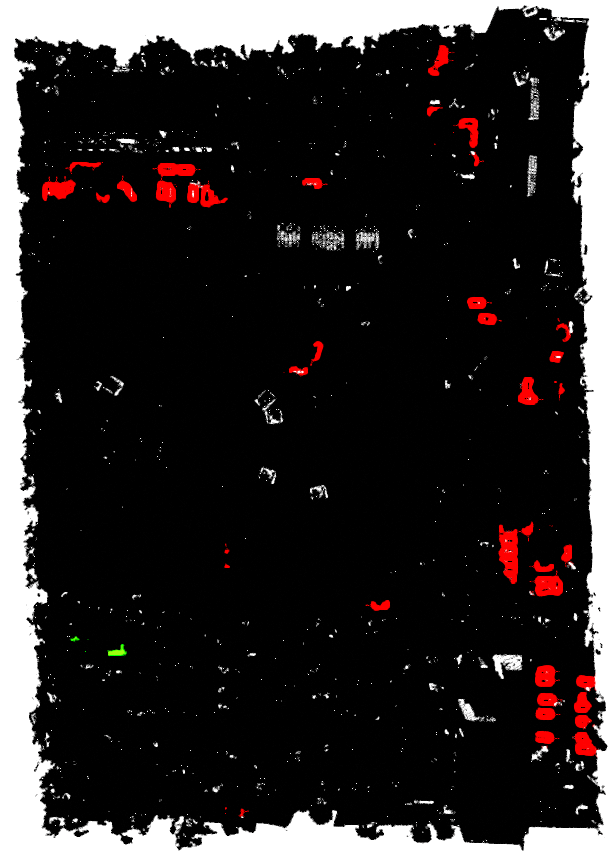}
    }%
    \subfigure[Voxel-RCNN]{
        \includegraphics[width=0.18\textwidth]{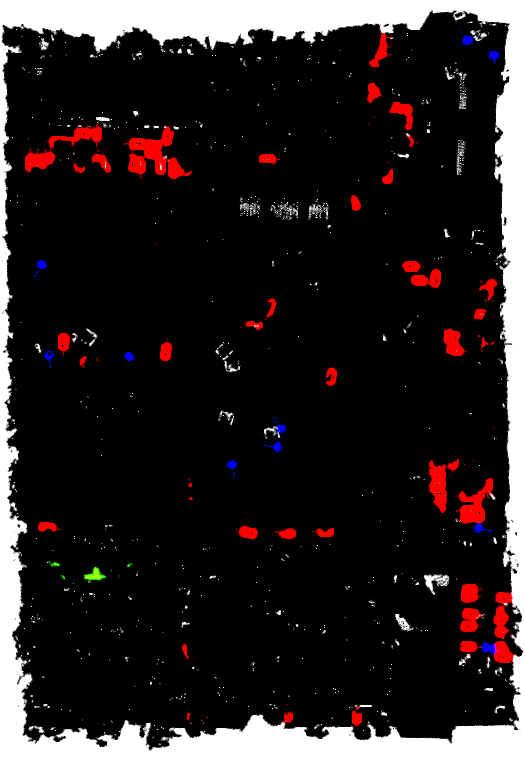}
    }%
    \subfigure[TransFusion-L]{
        \includegraphics[width=0.18\textwidth]{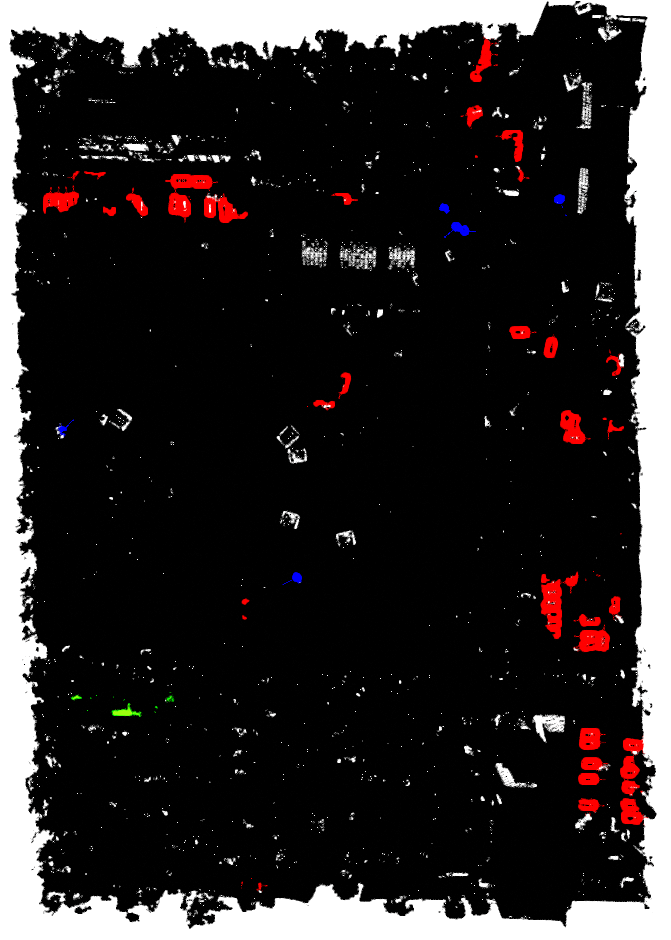}
    }%
    \subfigure[VoxelNext]{
        \includegraphics[width=0.18\textwidth]{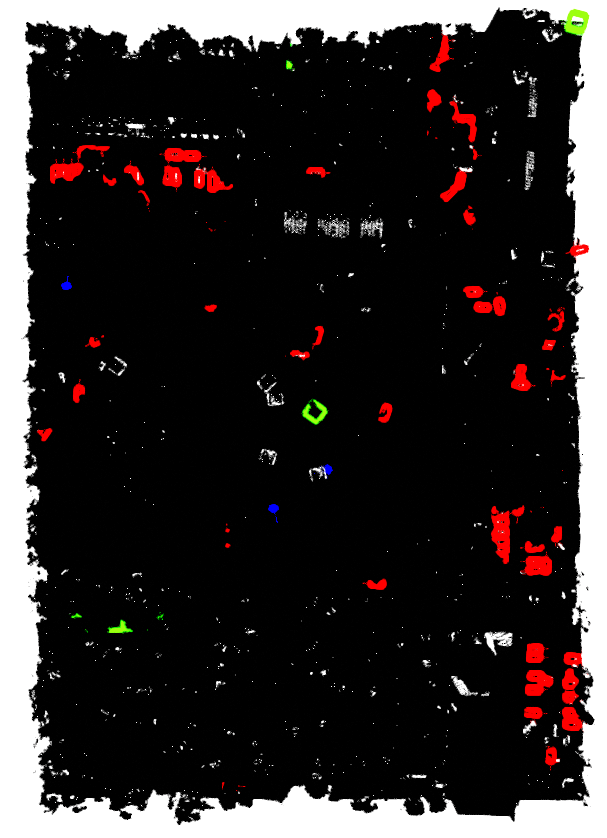}
    }%
    \subfigure[Proposed]{
        \includegraphics[width=0.18\textwidth]{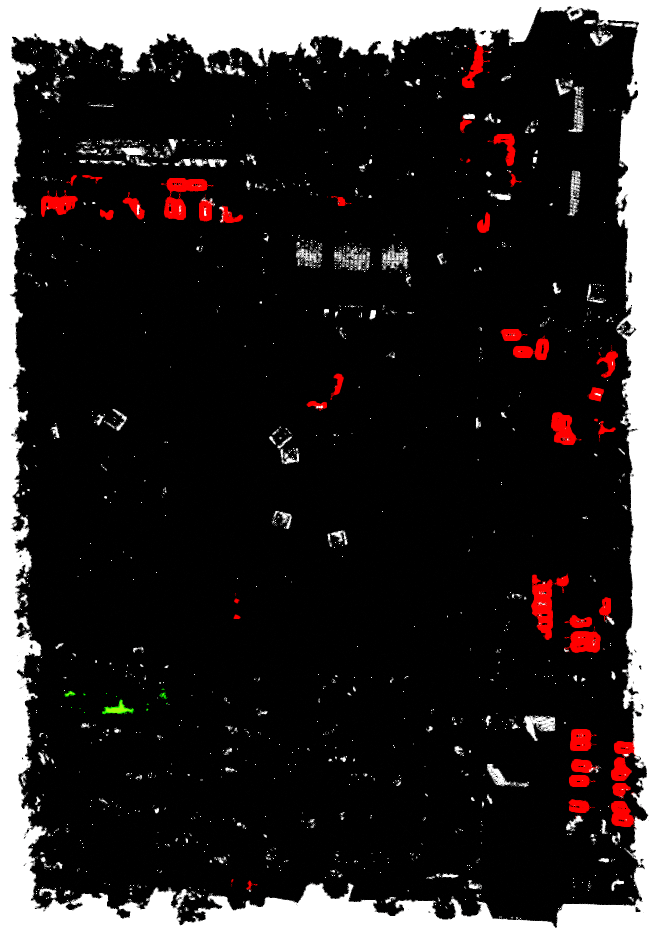}
    }%
    \caption{Visualization of the predicted 3D bounding boxes in HIT Campus dataset scene 2, where red represents vehicles, green represents tents, and blue represents boxes}
\end{figure}

\textbf{3) Performance on HIT-L2 test set:} To further verify the generalization ability of PiV-AHPC, we evaluate various methods on the HIT-L2 dataset, as shown in Table 5. Experiments show that PiV-AHPC achieves the best performance across all detection metrics. Under easy, moderate, and hard difficulty levels, PiV-AHPC's detection accuracy on $\text{mA}{{\text{P}}_{\text{BEV}}}$ exceeds the best comparison methods by 1.19\%, 2.07\%, and 1.98\% respectively. This performance improvement stems from the dual-branch structure precisely extracting heterogeneous HPCs features, combined with multi-level feature fusion enhancing information integration, significantly improving the network's adaptability to HPCs. Although PiV-AHPC performs best on the HIT-L2 dataset, its advantage over other methods is less pronounced than on the HIT-Campus dataset. This phenomenon may result from the smaller dataset size limiting PiV-AHPC's potential in complex scenes. Additionally, limited by 3D reconstruction accuracy, the pseudo point cloud assignment strategy may introduce significant spectral bias, interfering with detector learning.

\begin{table}
    \setlength{\tabcolsep}{4pt}
    \caption{PERFORMANCE OF 3D OBJECT DETECTION METHODS ON HIT CAMPUS TEST SET. THE (VEHICLE, TENT, BOX) RESULTS ARE REPORTED BY THE AP WITH (0.7,0.5,0.5) IOU THRESHOLD AND 40 RECALL POINTS.}
    \centering
    \footnotesize
    \begin{tabular}{c|c|c|c|c|c|c|c|c|c|c|c|c|c}
    \hline
        \multirow{2}{*}{Method} & \multirow{2}{*}{Difficulty} & \multicolumn{3}{c|}{Vehicle} & \multicolumn{3}{c|}{Tent} & \multicolumn{3}{c|}{Box} & \multicolumn{3}{c}{mAP} \\ \cline{3-14}
        & & BEV & 3D & AOS & BEV & 3D & AOS & BEV & 3D & AOS & BEV & 3D & AOS \\ \hline
        \multirow{3}{*}{PartA2} & Easy & 70.84 & 42.57 & 70.77 & 90.48 & 70.67 & 89.8 & 55.66 & 20.06 & 41.92 & 74.23 & 45 & 69.35 \\ \cline{2-14}
        & Mod & 59.97 & 29.01 & 59.63 & 87.24 & 63.6 & 82.79 & 43.43 & 13.62 & 32.87 & 63.65 & 35.88 & 58.5 \\ \cline{2-14}
        & Hard & 46.94 & 21.33 & 46.48 & 51.96 & 37.33 & 48.81 & 31.59 & 9.13 & 23.52 & 43.19 & 21.3 & 39.17 \\ \hline
        \multirow{3}{*}{CenterPoint} & Easy & 61.84 & 30.55 & 61.8 & 54.94 & 0.92 & 20.33 & 21.97 & 4.14 & 8.46 & 46.25 & 11.87 & 30.19 \\ \cline{2-14}
        & Mod & 52.88 & 22.02 & 52.66 & 73.36 & 7.89 & 35.14 & 24.3 & 2.67 & 10.8 & 50.18 & 10.86 & 32.87 \\ \cline{2-14}
        & Hard & 40.09 & 16 & 39.76 & 40.182 & 3.92 & 19.2 & 17.76 & 1.84 & 7.78 & 32.69 & 7.25 & 22.24 \\ \hline
        \multirow{3}{*}{PillarNet} & Easy & 51.12 & 12.48 & 49.77 & 81.1 & 78.37 & 78.94 & 14.62 & 4.32 & 13.18 & 49.33 & 31.73 & 47.6 \\ \cline{2-14}
        & Mod & 43.26 & 9.35 & 41.6 & 72.65 & 63.49 & 67.75 & 23.26 & 7.34 & 18.95 & 46.85 & 26.77 & 43 \\ \cline{2-14}
        & Hard & 34.8 & 7.03 & 32.93 & 39.94 & 36.36 & 37.16 & 20.13 & 6.5 & 16.39 & 30.97 & 15.28 & 28 \\ \hline
        \multirow{3}{*}{PV-RCNN++} & Easy & 66.64 & 40.13 & 66.5 & 76.62 & 15.86 & 56.22 & 41.32 & 9.56 & 27.11 & 61.52 & 21.85 & 49.94 \\ \cline{2-14}
        & Mod & 55.54 & 27.03 & 55.31 & 72.97 & 19.7 & 51.3 & 42.76 & 6.84 & 28.56 & 57.09 & 17.85 & 45.05 \\ \cline{2-14}
        & Hard & 41.74 & 18.68 & 41.41 & 37.56 & 9.93 & 26.05 & 30.184 & 4.15 & 19.83 & 36.52 & 10.92 & 29.1 \\ \hline
        \multirow{3}{*}{Voxel-RCNN} & Easy & 72.88 & 45.23 & 72.67 & 42.12 & 35.16 & 41.78 & 45.98 & 23.58 & 37.93 & 54.05 & 33.8 & 51.17 \\ \cline{2-14}
        & Mod & 67.81 & 33.09 & 66.84 & 51.77 & 45.96 & 49.27 & 51.66 & 17.38 & 42.72 & 57.88 & 31.76 & 53.58 \\ \cline{2-14}
        & Hard & 53.68 & 24.11 & 52.57 & 43.44 & 31.39 & 40.6 & 34.94 & 11.68 & 28.8 & 42.24 & 21.01 & 38.91 \\ \hline
        \multirow{3}{*}{TransFusion-L} & Easy & 64.42 & 27.09 & 64.33 & 58.87 & 5.42 & 36.68 & 13.99 & 1.82 & 6.13 & 45.76 & 11.44 & 35.71 \\ \cline{2-14}
        & Mod & 55.14 & 19.03 & 54.91 & 78.06 & 10.88 & 41.37 & 14.09 & 1.62 & 6.54 & 49.09 & 10.51 & 34.27 \\ \cline{2-14}
        & Hard & 43.21 & 14 & 42.82 & 47.83 & 5.84 & 24.5 & 11.66 & 1.09 & 5.44 & 34.24 & 6.98 & 24.25 \\ \hline
        \multirow{3}{*}{VoxelNext} & Easy & 69.14 & 39.01 & 69.04 & 88.11 & 17.83 & 41.15 & 19.08 & 5.19 & 10.11 & 58.78 & 20.68 & 40.1 \\ \cline{2-14}
        & Mod & 57.87 & 25.95 & 57.62 & 77.56 & 23.13 & 37.16 & 24.34 & 4.19 & 13.3 & 53.26 & 17.76 & 36.03 \\ \cline{2-14}
        & Hard & 44.05 & 18.78 & 43.74 & 53.62 & 14.15 & 25.35 & 18.68 & 3.15 & 9.96 & 38.78 & 12.03 & 26.35 \\ \hline
        \multirow{3}{*}{\textbf{PiV-AHPC}} & Easy & \textbf{73.33} & \textbf{48.79} & \textbf{72.36} & \textbf{99.22} & \textbf{98.69} & \textbf{98.89} & \textbf{66.13} & \textbf{37.64} & \textbf{59.55} & \textbf{79.56} & \textbf{61.71} & \textbf{76.93} \\ \cline{2-14}
        & Mod & 66.87 & \textbf{33.94} & 64.94 & \textbf{97.97} & \textbf{96.76} & \textbf{97.38} & \textbf{62.76} & \textbf{27.71} & \textbf{56.41} & \textbf{75.87} & \textbf{52.8} & \textbf{72.91} \\ \cline{2-14}
        & Hard & 52.97 & \textbf{24.62} & 50.85 & \textbf{72.5} & \textbf{60.54} & \textbf{71.39} & \textbf{49.04} & \textbf{21.03} & \textbf{43.91} & \textbf{58.17} & \textbf{35.39} & \textbf{55.38} \\ \hline
    \end{tabular}
\end{table}

\vspace{-10pt} 

Comparing the detection results from the three datasets reveals that many networks exhibit varying adaptability across different datasets, with challenges in maintaining consistent performance across diverse scenes and sensor types. In contrast, PiV-AHPC achieves outstanding results in all three datasets which have clearly different features, showcasing its exceptional generalization ability.

\begin{figure}[!htbp]
    \centering
    \subfigure[Ground Truth]{
        \includegraphics[width=0.17\textwidth]{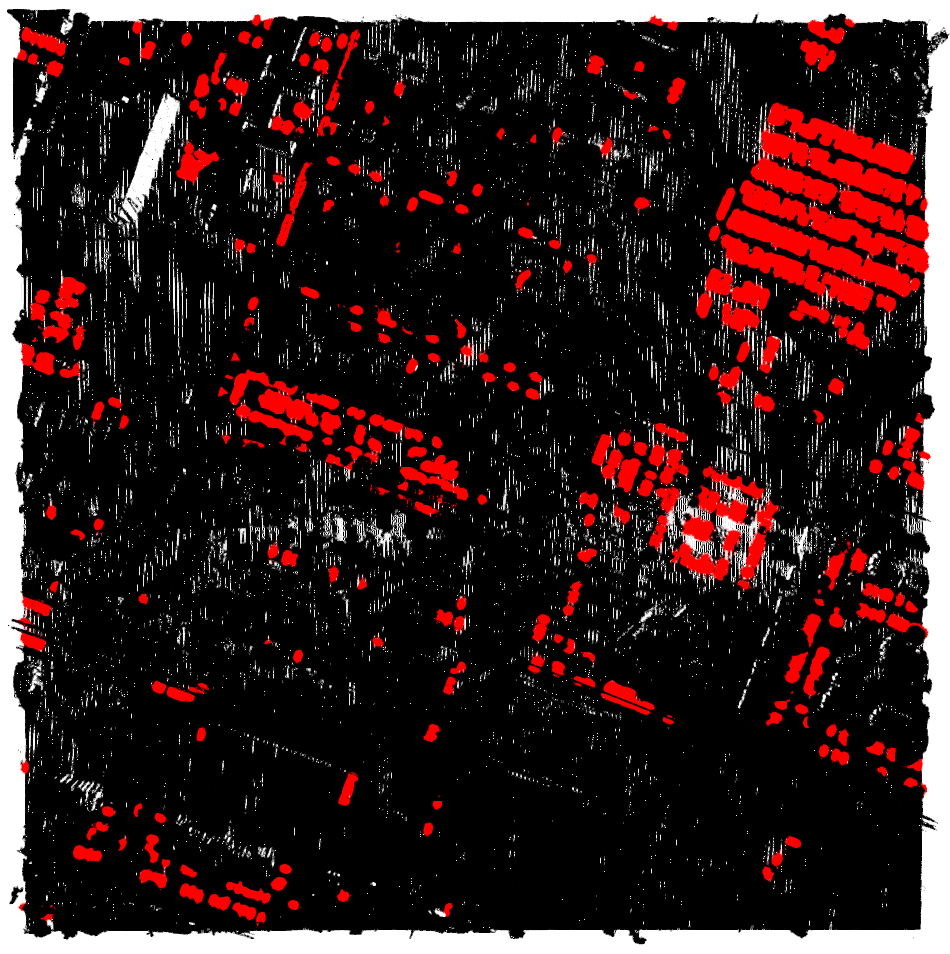}
    }%
    \subfigure[PartA2]{
        \includegraphics[width=0.17\textwidth]{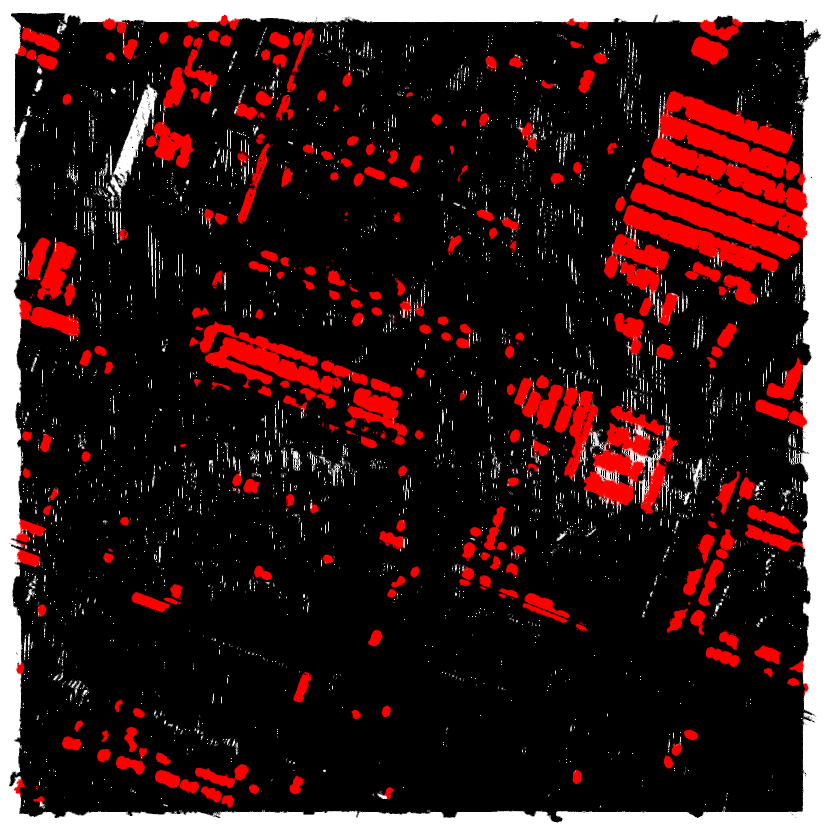}
    }%
    \subfigure[CenterPoint]{
        \includegraphics[width=0.17\textwidth]{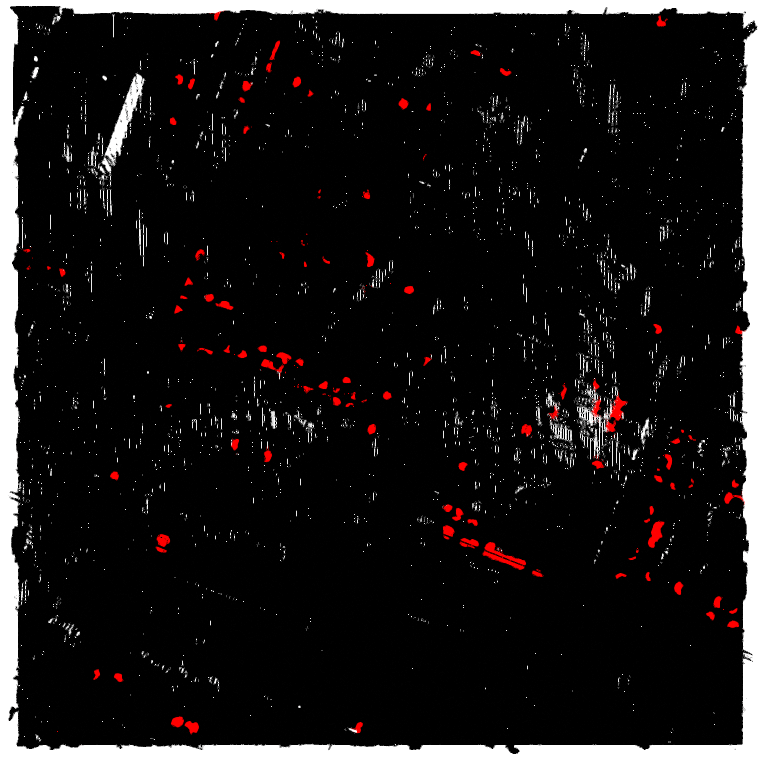}
    }%
    \subfigure[PillarNet]{
        \includegraphics[width=0.17\textwidth]{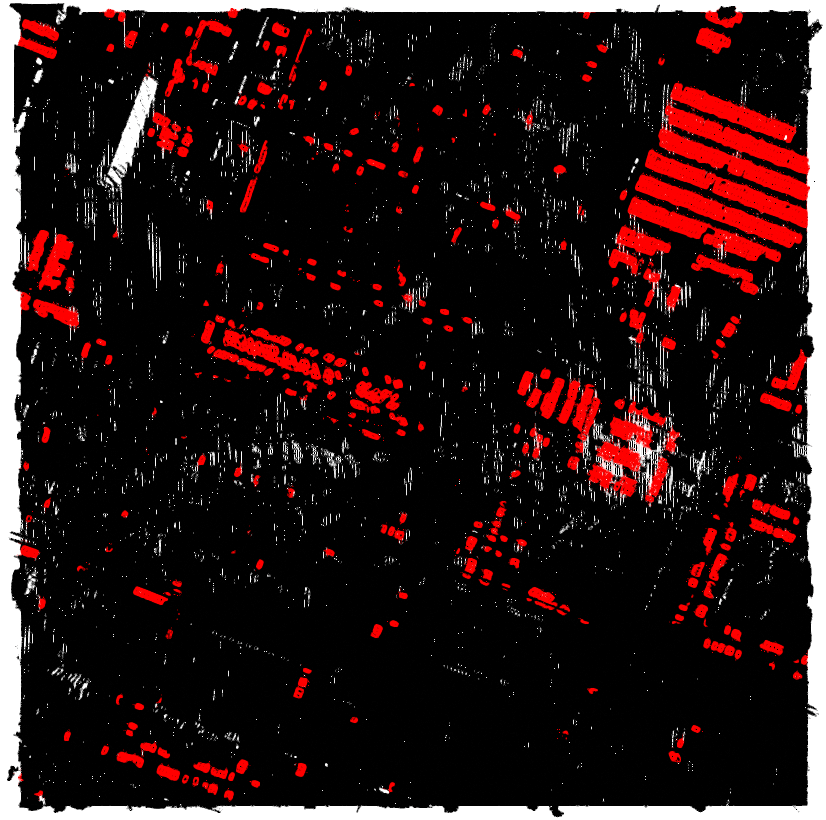}
    }%
    
    \subfigure[PV-RCNN++]{
        \includegraphics[width=0.17\textwidth]{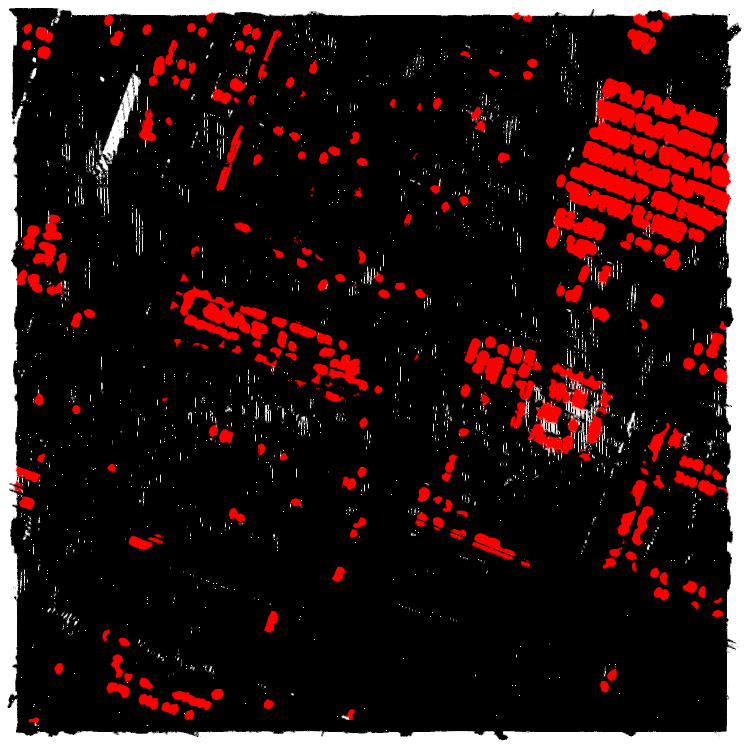}
    }%
    \subfigure[Voxel-RCNN]{
        \includegraphics[width=0.17\textwidth]{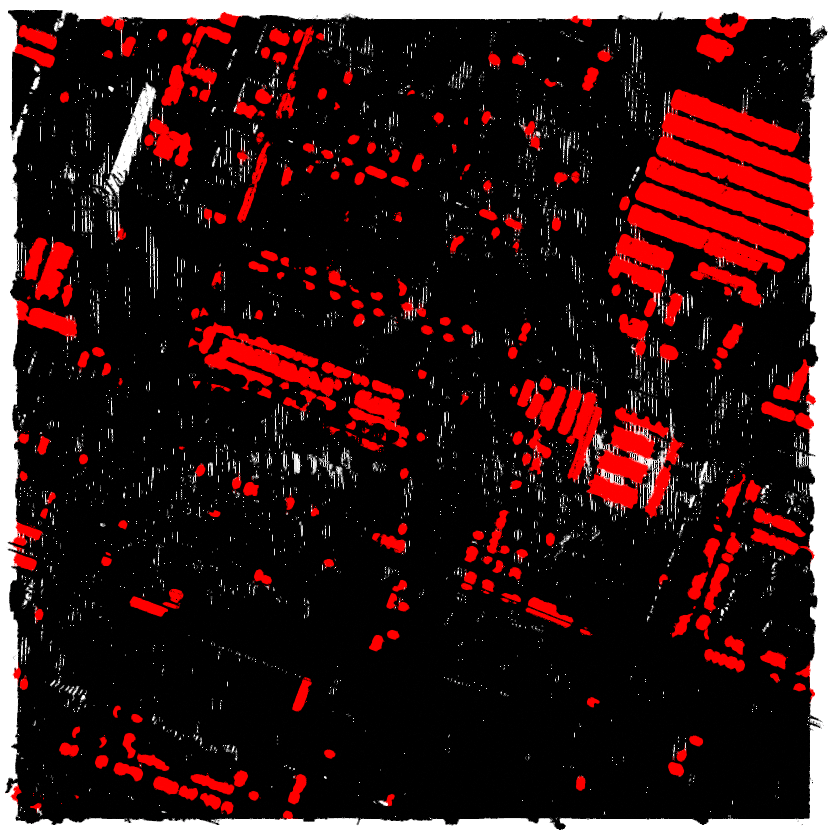}
    }%
    \subfigure[TransFusion-L]{
        \includegraphics[width=0.17\textwidth]{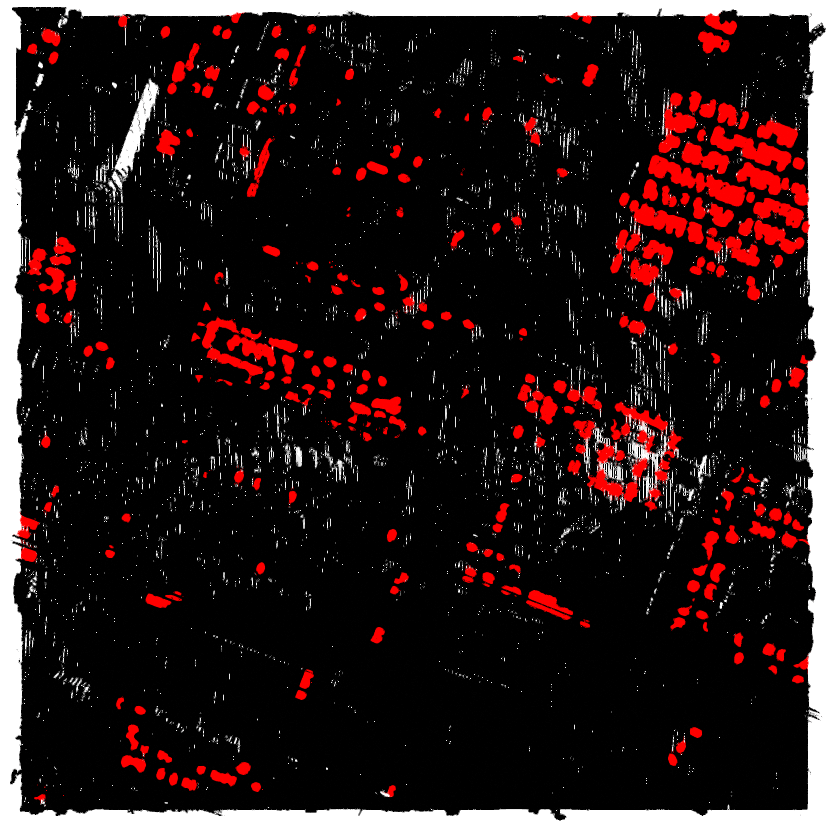}
    }%
    \subfigure[VoxelNext]{
        \includegraphics[width=0.17\textwidth]{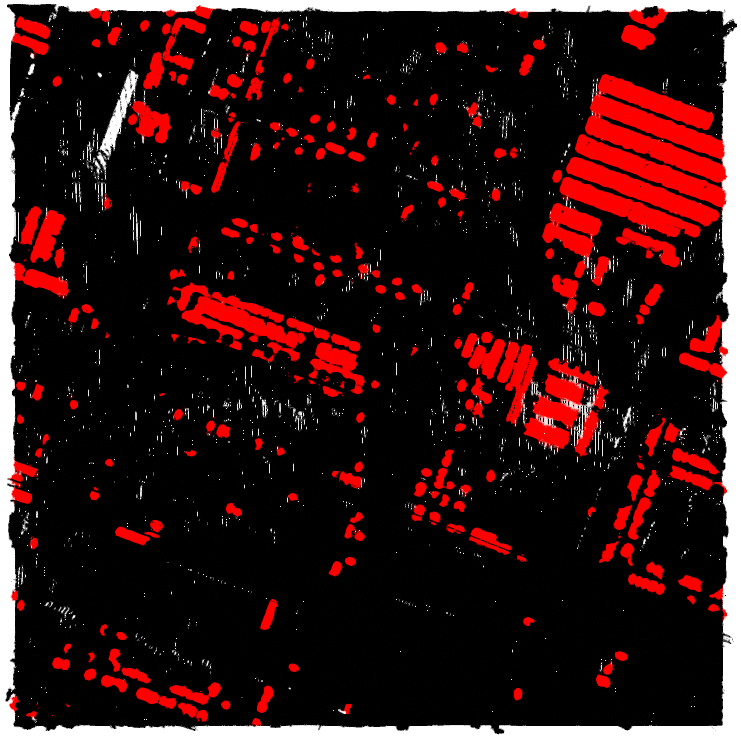}
    }%
    \subfigure[Proposed]{
        \includegraphics[width=0.17\textwidth]{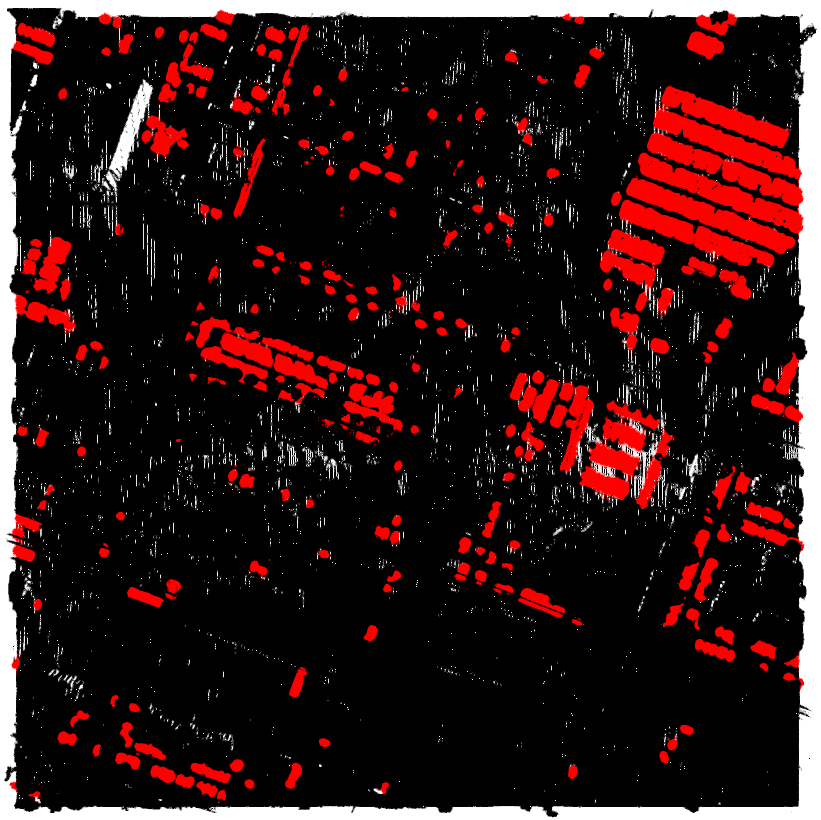}
    }%
    \caption{Visualization of the predicted 3D bounding boxes in Houston2018 dataset, where red represents vehicles}
\end{figure}


\vspace{-5pt}  

\begin{table}[!htbp]
    \caption{PERFORMANCE OF 3D OBJECT DETECTION METHODS ON HOUSTON2018 TEST SET. THE RESULTS ARE REPORTED BY THE AP WITH 0.7 AND 0.5:0.95 IOU THRESHOLD AND 40 RECALL POINTS.}
    \setlength{\tabcolsep}{5pt}
    \centering
    \footnotesize
    \begin{tabular}{c|c|c|c|c|c|c|c|c|c|c|c|c}
    \hline
        \multirow{2}{*}{Method} & \multicolumn{4}{c|}{$\text{A}{{\text{P}}_{\text{BEV}}}$} & \multicolumn{4}{c|}{$\text{A}{{\text{P}}_{\text{3D}}}$} & \multicolumn{4}{c}{AOS}\\ \cline{2-13}
        & Easy & Mod & Hard & 0.5:0.95 & Easy & Mod & Hard & 0.5:0.95 & Easy & Mod & Hard & 0.5:0.95 \\ \hline
        PartA2 & 64.48 & 69.78 & 67.67 & 47.48 & 50.81 & 55.33 & 50.33 & 35.78 & 64.10 & 69.30 & 66.95 & 47.22 \\ \hline
        CenterPoint & 45.33 & 55.16 & 64.61 & 31.76 & 26.53 & 33.45 & 42.64 & 21.82 & 45.22 & 54.79 & 63.95 & 31.69 \\ \hline
        PillarNet & 18.89 & 25.76 & 26.74 & 15.19 & 0.34 & 0.32 & 0.187 & 3.55 & 9.51 & 12.71 & 13.43 & 7.60 \\ \hline
        PV-RCNN++ & 74.39 & 73.21 & 67.13 & 53.53 & 58.01 & 56.04 & 50.11 & 40.12 & 74.34 & 72.98 & 66.69 & 53.50 \\ \hline
        Voxel-RCNN & 46.97 & 56.14 & 58.72 & 34.34 & 37.01 & 45.38 & 44.56 & 26.01 & 46.57 & 55.70 & 58.05 & 34.08 \\ \hline
        TransFusion-L & 80.38 & 82.81 & 77.34 & 57.22 & 52.53 & 53.65 & 52.10 & 39.85 & 80.32 & 82.56 & 76.78 & 57.18 \\ \hline
        VoxelNext & 59.95 & 67.55 & 73.19 & 42.72 & 31.71 & 39.61 & 47.22 & 25.30 & 59.91 & 67.26 & 72.49 & 42.69 \\ \hline
        \textbf{PiV-AHPC} & \textbf{82.70} & \textbf{85.89} & \textbf{79.62} & \textbf{61.44} & \textbf{63.21} & \textbf{69.19} & \textbf{62.58} & \textbf{45.63} & \textbf{81.19} & \textbf{84.68} & \textbf{78.22} & \textbf{60.40} \\ \hline
    \end{tabular}
\end{table}

\vspace{-10pt}  

\noindent

\subsection{Comparative Analysis of Different Occlusion Levels}
TABLE 6,7 display the object detection accuracy of PiV-AHPC under different levels of occlusion in two datasets. The HIT Campus Dataset does not include part occlusion tents, as well as occlusion-free tents and boxes. According to the table content, the algorithm’s detection accuracy shows a low correlation with the occlusion level, suggesting that occlusion has little impact on PiV-AHPC's performance. Specifically, in the HIT Campus Dataset, compared to part occlusion and occlusion-free scenes, the algorithm in high occlusion scene for vehicle objects achieves improvements of 1.75\% and 5.23\% in $\text{A}{{\text{P}}_{\text{BEV}}}$, and 3.76\% and 11.82\% in $\text{A}{{\text{P}}_{\text{3D}}}$. For part occlusion box objects, improvements in $\text{A}{{\text{P}}_{\text{BEV}}}$ and $\text{A}{{\text{P}}_{\text{3D}}}$ compared to high occlusion objects are 1.74\% and 2.14\%, respectively. The algorithm achieves the best detection accuracy for high occlusion objects in the Houston2018 dataset, with $\text{A}{{\text{P}}_{\text{3D}}}$ surpassing the other two types of objects by 2.56\% and 1.18\%, respectively. Spectral assignments of low-level objects in HPCs are often incorrect when occlusion occurs. However, PiV-AHPC demonstrates promising performance in high-occlusion scenes across two datasets. On the one hand, these results highlight the advantages of LiDAR, including its ability to penetrate occlusions and achieve 3D imaging. On the other hand, they demonstrate the effectiveness of PiV-AHPC's pillar branch, which has a sufficient vertical receptive field. This branch selectively utilizes spectral information based on the distribution of specific objects, suppressing interference from incorrect spectral values.

\begin{table}[!htbp]
    \centering
    \footnotesize
    
    \setlength{\tabcolsep}{5pt}
    \caption{PERFORMANCE OF 3D OBJECT DETECTION METHODS ON HIT-L2 TEST SET. THE (VEHICLE, CANOPY, TENT) RESULTS ARE REPORTED BY THE AP WITH (0.7,0.5,0.5) IOU THRESHOLD AND 40 RECALL POINTS.}
    \begin{tabular}{c|c|c|c|c|c|c|c|c|c}
    \hline
        \multirow{2}{*}{Method} & \multicolumn{3}{c|}{$\text{mA}{{\text{P}}_{\text{BEV}}}$} & \multicolumn{3}{c|}{$\text{mA}{{\text{P}}_{\text{3D}}}$} & \multicolumn{3}{c}{AOS}\\ \cline{2-10}
        & Easy & Mod & Hard & Easy & Mod & Hard & Easy & Mod & Hard \\ \hline
        PartA2 & 70.88 & 68.32 & 65.26 & 46.22 & 44.45 & 43.19 & 51.37 & 50.31 & 49.07 \\ \hline
        CenterPoint & 78.7 & 76.37 & 73.21 & 25.29 & 23.89 & 21.96 & 49.95 & 47.63 & 44.47 \\ \hline
        PillarNet & 18.31 & 17.83 & 17.3 & 6.32 & 6.16 & 6.13 & 12.37 & 12.02 & 11.67 \\ \hline
        PV-RCNN++ & 63.09 & 60.62 & 57.32 & 52.64 & 50.43 & 48.18 & 54.23 & 51.76 & 48.44 \\ \hline
        Voxel-RCNN & 77.23 & 74.83 & 71.7 & 57 & 55.56 & 53.77 & 62.85 & 61.12 & 59.12 \\ \hline
        TransFusion-L & 56.5 & 56.17 & 56.11 & 5.75 & 5.74 & 5.74 & 28.34 & 28.18 & 28.15 \\ \hline
        VoxelNext & 78.93 & 74.84 & 71.56 & 41.47 & 38.63 & 36.36 & 53.52 & 49.75 & 46.48 \\ \hline
        \textbf{PiV-AHPC} & \textbf{80.12} & \textbf{78.44} & \textbf{75.19} & \textbf{57.4} & \textbf{55.46} & \textbf{54.19} & \textbf{70.57} & \textbf{68.85} & \textbf{65.77} \\ \hline
    \end{tabular}
    
\end{table}

\begin{table}[!htbp]
    \centering
    \footnotesize
    \caption{Performance of PiV-AHPC on the HIT-Campus test set under different occlusion levels. "Free," "Part," and "High" refer to "Occlusion-Free," "Part Occlusion," and "High Occlusion"}
    \begin{tabular}{c|c|c|c|c|c|c|c|c|c}
    \hline
        \multirow{2}{*}{Metric} & \multicolumn{3}{c|}{Vehicle} & \multicolumn{2}{c|}{Box} & Tent & \multicolumn{3}{c}{mAP} \\ \cline{2-10}
        ~ & Free & Part & High & Part & High & High & Free & Part & High \\ \hline
        $\text{A}{{\text{P}}_{\text{BEV}}}$ & 63.35 & 66.83 & 68.58 & 53.09 & 51.35 & 72.5 & 63.35 & 59.96 & 64.14 \\ \hline
        $\text{A}{{\text{P}}_{\text{3D}}}$ & 50.12 & 58.18 & 61.94 & 22.19 & 20.05 & 60.54 & 50.12 & 40.19 & 47.51 \\ \hline
        AOS & 60.185 & 63.78 & 64.99 & 46.83 & 45.83 & 71.39 & 60.185 & 54.81 & 60.74 \\ \hline
    \end{tabular}
\end{table}

\begin{table}[!htbp]
    \centering
    \footnotesize
    \caption{Performance of PiV-AHPC on the Houston2018 test set under different occlusion levels. "Free," "Part," and "High" refer to "Occlusion-Free," "Part Occlusion," and "High OcclusionN”}
    \begin{tabular}{c|c|c|c|c|c|c|c|c}
    \hline
        \multicolumn{3}{c|}{Free} & \multicolumn{3}{c|}{Part} & \multicolumn{3}{c}{High} \\ \hline
        $\text{A}{{\text{P}}_{\text{BEV}}}$ & $\text{A}{{\text{P}}_{\text{3D}}}$ & AOS & $\text{A}{{\text{P}}_{\text{BEV}}}$ & $\text{A}{{\text{P}}_{\text{3D}}}$ & AOS & $\text{A}{{\text{P}}_{\text{BEV}}}$ & $\text{A}{{\text{P}}_{\text{3D}}}$ & AOS \\ \hline
        85.94 & 81.88 & 83.85 & 85.66 & 83.26 & 83.7 & 86.9 & 84.44 & 85.05 \\ \hline
    \end{tabular}
\end{table}

\subsection{Analysis of algorithm transferability}
Although PiV-AHPC utilizes the detector proposed by Voxel-RCNN, its core innovation lies in the feature extraction strategy tailored for HPCs, which we named PiV. Specifically, PiV utilizes a dual-branch encoder for pillar-voxel feature extraction combined with a multi-level feature fusion mechanism to effectively capture the characteristics of HPCs and generate a dense BEV feature map of the scene. Consequently, for dense or semi-sparse detectors, PiV can serve as the backbone network, directly producing dense BEV features for the detection head. Meanwhile, for fully sparse detection architectures, PiV can be adapted by removing the patch-wise feature fusion module to align with the sparse detection head. In this section, the transferability of the proposed method is demonstrated using two networks: CenterPoint based on center prediction and PV-RCNN++ based on point-voxel. Their $\text{mA}{{\text{P}}_{\text{3D}}}$ performance of the HIT Campus Dataset is illustrated in Figure 12. Compared to the original networks, the new algorithm with the transferred PiV strategy significantly improves the accuracy of all detection metrics. Specifically, for CenterPoint, the $\text{mA}{{\text{P}}_{\text{3D}}}$ improves by 6.31\%, 4.46\%, and 2.76\% across three difficulty levels, while for PV-RCNN++, the improvement is 20.66\%, 19.65\%, and 12.32\%. In contrast, the PiV-AHPC proposed in this article achieves improvements of 27.91\%, 21.04\%, and 14.38\% compared to the original algorithm. These experiments demonstrate the flexible transferability of the PiV strategy, which can be applied to various 3D object detection architectures and effectively enhance their detection capabilities for HPCs data. Furthermore, this strategy exhibits stronger applicability to two-stage detection models, as the rich information contained in the intermediate features of the dual branches can be further leveraged in the proposal refinement stage to improve the details of object classification and localization.

\begin{figure}[!htbp]
    \centering
    \includegraphics[width=0.6\textwidth,height=0.3\textwidth]{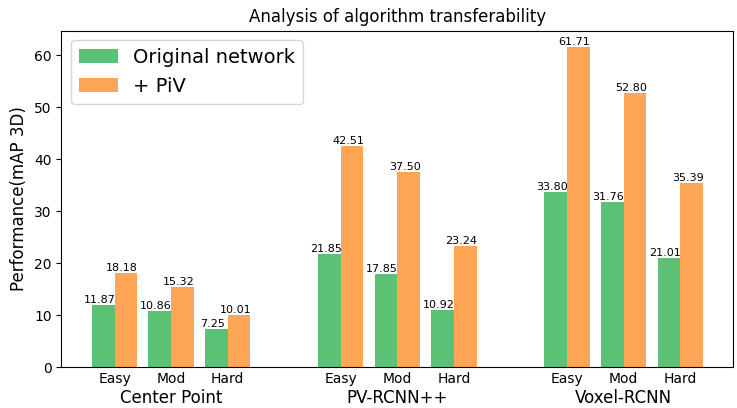}
    \caption{Performance of PiV on different 3D detector architectures for HIT Campus Testset}
    \end{figure} 

\begin{figure}[!htbp]
    \centering
    \includegraphics[width=1\textwidth]{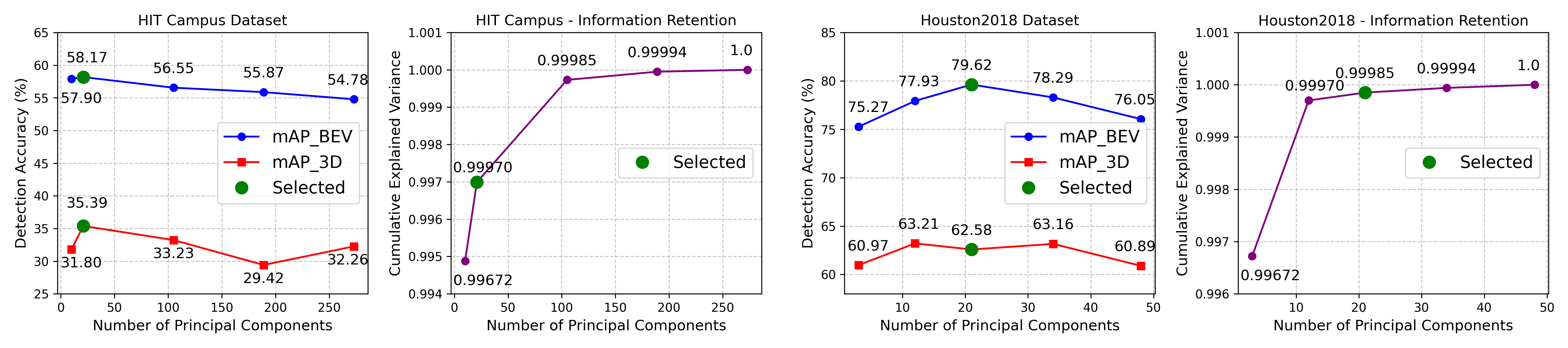}
    \caption{Performance of PiV-AHPC and cumulative explained variance under different numbers of principal components. For the HIT Campus dataset, we compare the results using 10, 21, 105, 189 principal components, and the original data. For the Houston2018 dataset, we compare 3, 12, 21, 34 principal components, and the original data.}
    \end{figure} 

\subsection{Ablation Study}
\textbf{1)Principal Component Selection:} We conducted ablation experiments on both HIT Campus and Houston2018 datasets, comparing detection performance under different numbers of principal components and their cumulative explained variance (reflecting information retention), as shown in Figure 13. For the HIT Campus dataset, when reduced to 21 principal components, the model achieved optimal detection performance under 'Hard' level, with $\text{mA}{{\text{P}}_{\text{3D}}}$ reaching 35.39\%, significantly outperforming the original data's 32.26\% and 10 principal components' 31.8\%. This indicates that excessive dimensions may contain redundant information, while over-reduction can lose fine-grained information, leading to decreased detection accuracy. 21 principal components both retain critical information (cumulative explained variance of 99.97\%) and reduce computational resource consumption. For the Houston2018 dataset, data with 21 principal components performed best on the $\text{A}{{\text{P}}_{\text{BEV}}}$ metric, although its $\text{A}{{\text{P}}_{\text{3D}}}$ was slightly lower than those with 12 and 34 dimensions. However, considering consistency with the HIT Campus dataset configuration, we chose 21 dimensions as the unified setting. Appropriate PCA dimensionality reduction not only improves computational efficiency but also enhances detection performance by reducing redundant information, achieving a balance between detection accuracy and computational cost.

\textbf{2)Effectiveness of pillar-voxel Encoder:} As the starting point of the network, the pillar-voxel dual-branch encoder is the foundation of the proposed method in this work. To test its effectiveness, we employed Voxel-RCNN as a baseline for single-branch networks and sequentially added the pillar branch, the sparse feature fusion module. TABLE 8 presents the mAP scores of each algorithm. The results indicate that our dual-branch encoder significantly enhances the detection capabilities of the original algorithm under three difficulty conditions. Taking the ‘Hard’ level of BEV, 3D, and angle metrics as examples: the initial precision of the vanilla network was 42.24\%, 21.01\%, and 38.91\%, respectively. After introducing the pillar branch, these metrics improved by 5.21\%, 1.82\%, and 5.52\%, respectively. Further addition of the sparse feature fusion module to facilitate information exchange between branches resulted in a continued improvement in detection accuracy by 2.74\%, 4.52\%, and 2.46\%. These ablation experiment results demonstrate the significant advantage of our proposed dual-branch encoder in capturing the complex spatial and spectral relationships between HPCs data. And this advantage will be further amplified by establishing inter-branch correlations.

\begin{table}
    \centering
    \footnotesize
    \caption{PERFORMANCE OF PILLAR-VOXEL STRUCTURE ON HIT CAMPUS TEST SET.}
    \begin{tabular}{c|c|c|c|c|c|c|c|c|c}
    \hline
        \multirow{2}{*}{Method} & \multicolumn{3}{c|}{Easy} &  \multicolumn{3}{c|}{Mod} & \multicolumn{3}{c}{Hard} \\ \cline{2-10}
        ~ & $\text{mA}{{\text{P}}_{\text{BEV}}}$ & $\text{mA}{{\text{P}}_{\text{3D}}}$ & AOS & $\text{mA}{{\text{P}}_{\text{BEV}}}$ & $\text{mA}{{\text{P}}_{\text{3D}}}$ & AOS & $\text{mA}{{\text{P}}_{\text{BEV}}}$ & $\text{mA}{{\text{P}}_{\text{3D}}}$ & AOS \\ \hline
        base & 54.05 & 33.8 & 51.17 & 57.88 & 31.76 & 53.58 & 42.24 & 21.01 & 38.91 \\ \hline
        + Pillar & 68.05 & 39.9 & 64.43 & 63.08 & 34.17 & 59.61 & 47.45 & 22.83 & 44.43 \\ \hline
        +SFF & \textbf{72.57} & \textbf{53.54} & \textbf{69.68} & \textbf{67.78} & \textbf{44.18} & \textbf{63.98} & \textbf{50.19} & \textbf{27.35} & \textbf{46.89} \\ \hline
        \multicolumn{10}{c}{Pillar indicates pillar branch, SFF indicates sparse feature fusion.}\\ 
    \end{tabular}
\end{table}

\begin{table}
    \centering
    \footnotesize
    \caption{PERFORMANCE OF KEY MODULES ON HIT CAMPUS TEST SET.}
    \begin{tabular}{c|c|c|c|c|c}
    \hline
        P & A & W & $\text{mA}{{\text{P}}_{\text{BEV}}}$ & $\text{mA}{{\text{P}}_{\text{3D}}}$ & mAOS \\ \hline
        \checkmark & \checkmark & \checkmark & \textbf{58.17} & \textbf{35.39} & \textbf{55.38} \\ \hline
        \checkmark & ~ & ~ & 50.65({-7.52}) & 29.34({-6.05}) & 47.62({-7.76}) \\ \hline
        \checkmark & \checkmark & ~ & 53.98({-4.19}) & 32.05({-3.34}) & 50.94({-4.44}) \\ \hline
        ~ & \checkmark & \checkmark & 55.30({-3.13}) & 33.19({-2.2}) & 54.26({-1.12}) \\ \hline
    \end{tabular}
\end{table}

\textbf{3)Effectiveness of Key Modules:} In addition to the dual-branch encoder, we designed the patch-wise feature fusion (P) module as well as the multi-scale feature aggregation module with additional downsampling layers (A) and weighted elevation compression (W). We conducted ablation studies on these key modules, using the algorithm's detection accuracy at the ‘Hard’ level as the evaluation criterion, as shown in TABLE 9. It can be observed that the absence of any module leads to a decrease in detection capability to varying degrees. Specifically, removing the ‘P’ resulted in a 2.2\% decrease in $\text{mA}{{\text{P}}_{\text{3D}}}$, while discarding the 'W' module led to a 3.34\% decrease. Eliminating the 'A' module resulted in a significant accuracy drop to 29.34\%. These experiments illustrate the significant role of the proposed key modules. The multi-scale feature aggregation module, with its diverse 3D receptive fields and selective elevation compression, significantly enhanced model performance. Although the patch-wise feature fusion module had limited impact on accuracy, its Transformer architecture demonstrated potential for application on large-scale datasets, offering possibilities for further enhancing model performance.

\section{Conclusion}
In this work, we analyze the advantages and challenges of using HPCs in airborne scenes, and propose the first 3D object detection network, PiV-AHPC, designed for the airborne HPCs, filling a gap in this research field. Specifically, we develop the pillar-voxel dual-branch encoder to capture the heterogeneous 3D spatial-spectral features from HPCs and devise the multi-level feature fusion mechanism to enhance information exchange between branches, enabling adaptive feature selection and fusion. The experimental results verify that PiV-AHPC maintains the best detection performance on two HPCs datasets with significant feature differences and exhibits strong generalization ability. In addition, the model can be flexibly migrated to multiple detection architectures based on BEV prediction, which indicates the potential for further improvement of its detection accuracy. We hope our research could provide a novel solution for the task of 3D object detection in airborne HPCs.

Although PiV-AHPC demonstrates advantages in multiple aspects, there is still room for optimization in the following areas. Firstly, the current feature fusion approach is relatively straightforward, and future work could focus on designing more targeted modules to further explore the deep correlations between dual-branch features. Secondly, due to equipment limitations, experiments were conducted using registered HPCs. Further evaluation and optimization on real airborne hyperspectral LiDAR data are needed in the future.



\section*{Acknowledgements}
This work was supported by the National Science Fund for Distinguished Young Scholars under Grant 62025107 and by the Open Fund Project of KuiYuan Laboratory (Grant No. KY202423)

\endgroup
\end{document}